\documentclass[11pt]{article}

\usepackage[final]{acl}

\usepackage{times}
\usepackage{latexsym}

\usepackage[T1]{fontenc}
\usepackage[utf8]{inputenc}

\usepackage{microtype}

\usepackage{inconsolata}

\usepackage{graphicx}

\usepackage{booktabs}   % 테이블 선(toprule, midrule 등)
\usepackage{pifont}     % 체크/엑스 아이콘 (ding)
\usepackage{xcolor}     % 색상
\usepackage{adjustbox}  % 테이블 크기 자동 조절
\usepackage{amssymb}    % 삼각형 기호 등 수학 기호
\usepackage{tabularx}
\usepackage{enumitem}
\usepackage{amsmath}
\usepackage{multirow}
\usepackage{array}
\usepackage{bm}
\usepackage{kotex}
\usepackage{makecell}
\newcommand{\yes}{\textcolor{green!50!black}{\ding{51}}} % 체크 표시 (녹색)
\newcommand{\no}{\textcolor{red!40!black}{\ding{55}}}    % 엑스 표시 (적색)
\newcommand{\pmark}{\textcolor{orange!70!black}{\ensuremath{\triangle}}} % 세모 표시 (주황색)

\newcommand{\best}[1]{\textbf{#1}}
\newcommand{\second}[1]{\underline{#1}}
\newcommand{\worst}[1]{\textbf{\textcolor{red!70!black}{#1}}}
\newcommand{\secworst}[1]{\textcolor{red!70!black}{\underline{#1}}}
\newcommand{\cmark}{\ding{51}}

\title{Measuring the Depth of LLM Unlearning via Activation Patching}

\author{
\textbf{Jaeung Lee, Dohyun Kim, Jaemin Jo\thanks{Corresponding author}}\\
Sungkyunkwan University\\
Republic of Korea\\
\texttt{\{dlwodnd00, kimdoh0423, jmjo\}@skku.edu}
}

\begin{document}
\maketitle
\begin{abstract}
Large language model (LLM) unlearning has emerged as a crucial post-hoc mechanism for privacy protection and AI safety, yet auditing whether target knowledge is truly erased remains challenging.
Existing output-level metrics fail to detect when this knowledge remains recoverable from internal representations.
Recent white-box studies reveal such residual knowledge but often rely on auxiliary training or dataset-specific adaptations, leaving no generalizable metric.
We close this gap with the \textsc{Unlearning Depth Score} (\textsc{UDS}), a metric that quantifies the mechanistic depth of unlearning via activation patching.
\textsc{UDS} first identifies layers that encode the target knowledge using a retain model baseline, then measures how much of it is erased in the unlearned model on a 0--1 scale.
In a meta-evaluation across 20 metrics on 150 unlearned models spanning 8 methods, \textsc{UDS} achieves the highest faithfulness and robustness, confirming our causal approach as the most reliable for unlearning evaluation.
Case studies further show that \textsc{UDS} uncovers residual knowledge obscured from observational metrics by representational shifts, with erasure depth varying across prompt types.
% Case studies further reveal that white-box metrics can disagree at the layer level and that erasure depth varies across examples.
We provide guidelines for integrating \textsc{UDS} into existing benchmarking frameworks and streamlining the evaluation pipeline.%
% \footnote{Code and data will be made publicly available upon acceptance.}
\footnote{Code and data are available at \url{https://github.com/gnueaj/unlearning-depth-score}.}
\end{abstract}
\section{Introduction}

Large language models (LLMs) memorize substantial portions of their training data \citep{tirumala2022memorization}, posing risks to privacy and AI safety when such data includes sensitive personal information or hazardous knowledge \citep{carlini2021extracting, bengio2025aisafety}. 
LLM unlearning addresses this by removing target knowledge from a trained model while preserving its general capabilities \citep{jang2023knowledge}.
The goal is to produce a model indistinguishable from one trained entirely without the target data \citep{bourtoule2021machine}, and a growing body of methods now pursues this objective \citep[e.g.,][]{jang2023knowledge, zhang2024negative, li2024wmdp}. 
% \ju{이 문장 다음문단으로 보내고 다음문단 두문단으로 쪼개기? Yet, as a growing~~으로 시작}

Yet, a fundamental question remains: \emph{how do we verify that knowledge has been genuinely removed?}
Recent benchmarking efforts have sought to systematically evaluate existing methods \citep{maini2024tofu, shi2025muse, li2024wmdp} and visually compare them \citep{lee2026comparator}, advocating that a reliable metric should exhibit \emph{faithfulness} (accuracy in detecting knowledge) and \emph{robustness} (stability under interventions) \citep{dorna2025openunlearning}.
% Alongside these efforts, several studies have identified residual knowledge within ostensibly unlearned models through white-box analyses \citep{hong2024dissecting, lynch2024eight, guo2025mechanistic}.
% Since adversaries can restore this knowledge through lightweight fine-tuning \citep{lo2024relearn} or activation manipulation \citep{lynch2024eight}, evaluation must look beyond final outputs \ju{logits}. 
% However, current white-box approaches often require auxiliary training or are tied to specific datasets, leaving no generalizable score for systematic comparison.

While these benchmarking frameworks rely primarily on output-level metrics, adversaries can restore ostensibly erased knowledge through lightweight fine-tuning \citep{fan2025relearn} or activation manipulation \citep{seyitoglu2024extracting, jang2026suppression}.
This vulnerability suggests that evaluation must move beyond output logits, leading several studies to explore white-box analyses to identify residual knowledge \citep{hong2024dissecting, lynch2024eight, guo2025mechanistic}.
However, current white-box approaches often require auxiliary training or are tied to specific datasets, leaving no generalizable score for systematic method comparison.

To address these limitations, we propose the \textsc{Unlearning Depth Score} (\textsc{UDS}), a training-free, causal, and dataset-invariant metric that quantifies the mechanistic depth of unlearning via activation patching.
We define \emph{depth} as how far unlearning penetrates the model's internals, rather than merely altering final outputs. 
\textsc{UDS} operates in two stages: (1) a baselining stage that identifies knowledge-encoding layers by patching hidden states from the retain model (i.e., trained without target data) into the full model (i.e., trained on all data including the target), and (2) a quantification stage that patches the unlearned model's hidden states into the full model to measure how much of the encoded knowledge persists.
Unlike prior diagnostic analyses, \textsc{UDS} causally intervenes to test whether the knowledge is recoverable, yielding a per-example score from 0 (intact) to 1 (erased) that reflects the erasure depth across layers.

In our meta-evaluation of 20 metrics on 150 unlearned models spanning 8 methods, \textsc{UDS} achieves the highest faithfulness (AUC-ROC 0.971) and robustness (HM 0.932), outperforming output-level metrics and white-box baselines.
Our case studies demonstrate that causal evaluation uncovers residual knowledge obscured by representational shifts that mislead observational metrics, and that erasure depth varies across prompt types within a single method.
% \jj{이게 무엇을 imply하는지?}
% Finally, we provide guidelines for integrating \textsc{UDS} into existing frameworks and streamlining the evaluation pipeline.
Finally, we provide guidelines for integrating \textsc{UDS} into the privacy axis of existing frameworks and streamlining the evaluation pipeline.
% Our case studies further demonstrate that causal evaluation uncovers residual knowledge that observational metrics miss, and we provide guidelines for integrating UDS into existing frameworks and streamlining the evaluation pipeline.

To summarize, our contributions are:
\begin{itemize}[leftmargin=1.5em,topsep=1pt,itemsep=1pt]
\item \textsc{Unlearning Depth Score} (\textsc{UDS}), a metric that quantifies the mechanistic depth of unlearning via two-stage activation patching (\S\ref{sec:uds}).
\item A meta-evaluation of 20 metrics on 150 unlearned models over 8 methods, demonstrating that our causal approach most reliably evaluates knowledge erasure (\S\ref{sec:meta-eval}).
\item Case studies uncovering residual knowledge that observational metrics miss, with per-example analysis of erasure depth (\S\ref{sec:case-study}).
\item Guidelines for integrating \textsc{UDS} into the privacy axis of existing benchmarking frameworks and streamlining robustness evaluation (\S\ref{sec:implications}).
% \item Case studies uncovering residual knowledge that observational metrics miss, with per-example analysis of erasure depth across prompt types, and guidelines for integrating \textsc{UDS} into existing benchmarking frameworks.
% \item Case studies uncovering residual knowledge obscured by representational shifts that mislead observational metrics, with per-example analysis, and guidelines for integrating \textsc{UDS} into existing benchmarking frameworks.
% \item Case studies showing that \textsc{UDS} circumvents layer-level shifts misleading observational metrics, with per-example analysis and guidelines for integrating it into existing frameworks.
% \item Case studies uncovering layer-level metric disagreements and example-level variation in erasure depth within a single method, alongside guidelines for integrating \textsc{UDS} into existing frameworks.
\end{itemize}
\section{Background and Related Work}\label{sec:related}

\subsection{LLM Unlearning}
Given a trained model, a forget set $D_f$, and a retain set $D_r$, the goal of LLM unlearning is to produce a model indistinguishable from one trained only on $D_r$ \citep{bourtoule2021machine}, preserving general capabilities \citep{yao2024large}.

\paragraph{Methods.}
The simplest approach, gradient ascent \citep{jang2023knowledge}, maximizes loss on $D_f$, but unconstrained optimization leads to catastrophic collapse.
GradDiff \citep{yao2024large, maini2024tofu} mitigates this by jointly minimizing loss on $D_r$, though balancing opposing gradients remains fragile.
NPO \citep{zhang2024negative} reframes this tension through preference optimization, treating forget set completions as dispreferred, and SimNPO \citep{fan2025simplicity} simplifies this by removing the reference model and normalizing by response length.
IdkNLL and IdkDPO \citep{maini2024tofu} instead train the model to produce alternative responses (e.g., ``I don't know''), and AltPO \citep{mekala2025alternate} extends this with in-domain positive feedback on plausible alternatives.
RMU \citep{li2024wmdp} intervenes at the representation level, misdirecting hidden states toward random targets, while UNDIAL \citep{dong2025undial} uses self-distillation on adjusted logits to steer output distributions away from $D_f$.
Models unlearned with these methods across hyperparameter sweeps form the evaluation pool for our metric comparison in \S\ref{sec:meta-eval}.

\paragraph{Evaluation Frameworks.} 
% \jj{evaluation 보다는 evaluation Framework가 적당해보임.}
% \citep{shokri2017membership,shi2024detecting}
Unlearning is typically evaluated along three axes: \emph{memorization}, whether the model can still reproduce forget set knowledge; \emph{privacy}, whether an adversary can detect that the model was trained on $D_f$; and \emph{utility}, whether general performance is preserved.
Benchmarks such as TOFU \citep{maini2024tofu}, MUSE \citep{shi2025muse}, and WMDP \citep{li2024wmdp} addressed these concerns, and OpenUnlearning \citep{dorna2025openunlearning} consolidated them into a unified framework.
OpenUnlearning also introduced meta-evaluation to assess metric reliability itself by two criteria: \emph{faithfulness}, whether the metric can distinguish models with vs.\ without forget set knowledge, and \emph{robustness}, whether it remains stable under interventions like quantization and fine-tuning.
We extend this framework by introducing a symmetric robustness criterion. 
Unlike the original formulation that solely penalizes knowledge recovery, our symmetric formulation penalizes instability in either direction (see \S\ref{subsec:meta-setup}).

\subsection{White-box Evaluation of LLM Unlearning} 
% \jj{여기가 진짜 evaluation같은데 지금은 white-box ``analysis''로 되어있음. 대부분 score가 아니라서 이렇게 서술한 듯 한데, 느낌은 우리보다 더 원대한 것들이 있는 느낌. white-box analysis가 자주 쓰이는 말이긴 한데, evaluation 정도로 좁히는게 필요해 보임. }
Beyond output-level evaluation, a variety of techniques can probe model internals.
CKA \citep{kornblith2019similarity} compares representational geometry across layers, Logit Lens \citep{nostalgebraist2020logitlens} decodes intermediate hidden states through the model's prediction head, and Fisher Information \citep{kirkpatrick2017overcoming} quantifies parameter sensitivity to specific data. 
% and activation patching \citep{meng2022locating} causally tests knowledge by patching hidden states between models.
Activation patching \citep{meng2022locating} causally tests knowledge by substituting hidden states between runs of one model on different inputs; cross-model activation patching extends this to two models on the same input \citep{prakash2024finetuning}.

Applying these techniques to unlearning, several studies have shown that seemingly unlearned models preserve forget set knowledge internally.
\citet{xu2025unlearning} use CKA and Fisher diagnostics to characterize reversibility of unlearning across layers.
\citet{lynch2024eight} train probes on hidden states to detect latent knowledge invisible to output metrics.
\citet{guo2025mechanistic} use causal tracing to localize factual recall circuits, then confirm residual knowledge with trained probes.
\citet{hong2025intrinsic} project MLP value vectors into vocabulary space to show that parametric knowledge traces persist after unlearning.
\citet{patil2024can} apply logit lens projections to demonstrate that intermediate layers still decode supposedly erased knowledge.
\citet{hong2024dissecting} use parameter restoration to show that fine-tuning-based unlearning modifies MLP coefficient scores in the final layers without altering the value vectors, leaving stored knowledge intact.

These studies reveal residual knowledge, but they are primarily diagnostic: none provides a standardized, comparable score that generalizes across forget sets (Table~\ref{tab:related-work}).
\textsc{UDS} addresses this with a training-free, causal, dataset-invariant score for systematic method comparison. 
% \jj{이 문장 명쾌한데, 인트로 마지막 문단에 먼저 적혀야함}
% Requires: \usepackage{booktabs, pifont, xcolor, array}
\begin{table}[t]
\centering
\begin{adjustbox}{max width=\linewidth}
\begin{tabular}{@{}lcccc@{}}
\toprule
\textbf{Work}              & \textbf{Train-Free} & \textbf{Causal} & \textbf{Data-Inv} & \textbf{Score} \\
\midrule
\citet{xu2025unlearning}   & \yes & \no    & \yes & \no  \\
\citet{lynch2024eight}     & \no  & \no    & \yes & \no  \\
\citet{guo2025mechanistic} & \no  & \pmark & \no  & \no  \\
\citet{hong2025intrinsic}  & \yes & \pmark & \no  & \no  \\
\citet{patil2024can}       & \yes & \no    & \yes & \no  \\
\citet{hong2024dissecting} & \yes & \yes   & \no  & \yes \\
\midrule
\textbf{UDS (Ours)}        & \yes & \yes   & \yes & \yes \\
\bottomrule
\end{tabular}
\end{adjustbox}
\caption{Comparison of white-box unlearning analysis.
\textbf{Train-Free}: no auxiliary training.
\textbf{Causal}: evaluates knowledge causally (\pmark{} = causal localization but observational evaluation).
\textbf{Data-Inv}: directly applicable to new forget sets.
\textbf{Score}: proposes a metric quantifying residual forget set knowledge.
}
\label{tab:related-work}
\end{table}

\section{The \textsc{Unlearning Depth Score}}\label{sec:uds}
\begin{figure*}[t]
    \centering
    \includegraphics[width=\textwidth]{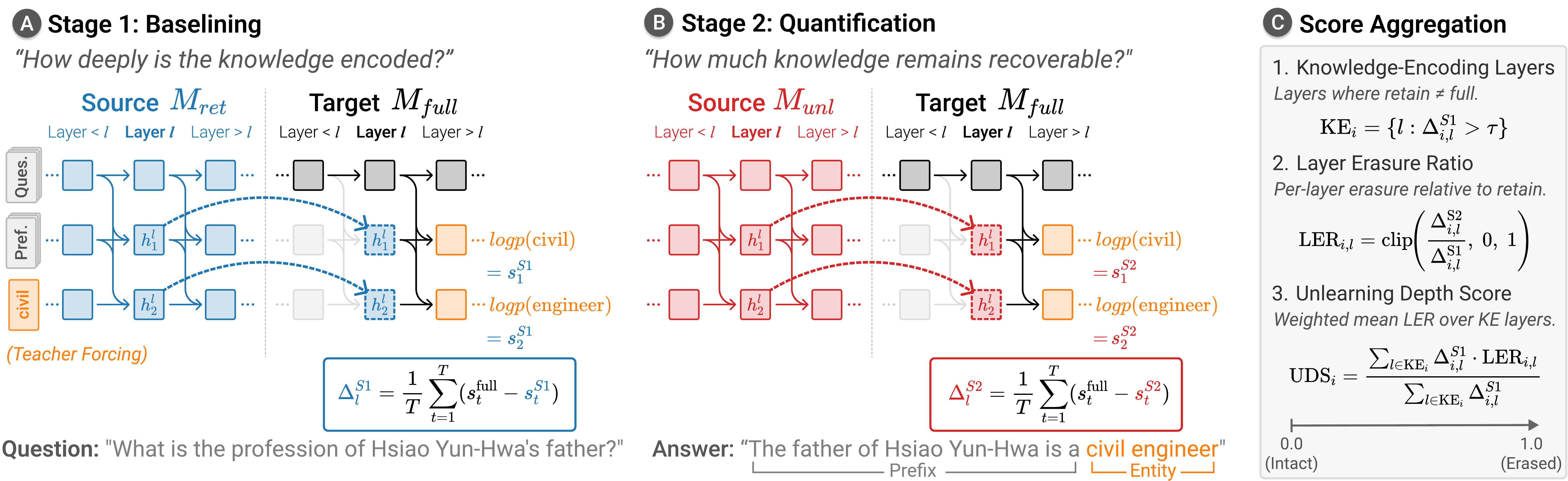}
\caption{Overview of \textsc{UDS} for a single forget set example.
\textbf{(A)}~Stage~1 patches hidden states from $M_{\text{ret}}$ into $M_{\text{full}}$ at each layer to measure how deeply the forget set knowledge is encoded.
\textbf{(B)}~Stage~2 repeats this with $M_{\text{unl}}$ as source to quantify how much encoded knowledge remains recoverable.
\textbf{(C)}~Stage~2 degradation is compared against Stage~1 at each layer to compute erasure ratios, which are weighted and aggregated into a single 0--1 score.}
    \label{fig:pipeline}
\end{figure*}
% \vspace{-20mm}
In this section, we describe \textsc{UDS}, a metric that quantifies the depth of unlearning by measuring how recoverable forget set knowledge is via activation patching (Figure~\ref{fig:pipeline}).
We define the problem setup (\S\ref{subsec:setup}), describe the patching procedure (\S\ref{subsec:patching}), discuss an efficient implementation strategy (\S\ref{subsec:efficient}), and validate the metric across model scales (\S\ref{subsec:scale}).
Appendix~\ref{app:ablation} provides further ablations to justify our design choices.
% In this section, we describe \textsc{UDS}, a metric that quantifies the depth of unlearning by measuring how recoverable forget set knowledge is via activation patching (Figure~\ref{fig:pipeline}).
% We define the problem setup (\S\ref{subsec:setup}), describe the patching procedure (\S\ref{subsec:patching}), and discuss an efficient implementation strategy (\S\ref{subsec:efficient}). Appendix~\ref{app:ablation} provides ablations to validate our design choices across model scales and prompt variations.

\subsection{Problem Setup}\label{subsec:setup}
\paragraph{Terminology.}
We consider three models:
(1)~$M_{\text{full}}$, trained on the full dataset $D_r \cup D_f$;
(2)~$M_{\text{ret}}$, trained only on $D_r$ (the gold standard for unlearning);
and (3)~$M_{\text{unl}}$, obtained by applying an unlearning method to $M_{\text{full}}$.
Following \citet{ghandeharioun2024patchscopes}, we call the model whose hidden states are extracted the \emph{source} and the model that receives them the \emph{target}.

\paragraph{Input and Measurement.}
Each forget set example $i$ consists of an input context $x_i$ and an entity span $y_i = (y_{i,1}, \dots, y_{i,T_i})$.
We focus on entity spans because common template phrases are predictable regardless of knowledge retention.
To avoid generation noise and enable stable per-token comparison across models, we use teacher forcing: the full input sequence is fed in a single forward pass.
The model predicts each entity token $y_{i,t}$ conditioned on the ground-truth prefix $y_{i,<t}$.
In autoregressive language models, the hidden state at position $p$ predicts token $p{+}1$; we therefore examine positions $y_{i,1}{-}1$ through $y_{i,T_i}{-}1$.

\subsection{Two-Stage Activation Patching}\label{subsec:patching} 
\textsc{UDS} proceeds in two stages: Stage~1 patches $M_{\text{ret}}$'s hidden states into $M_{\text{full}}$ to establish a baseline, and Stage~2 replaces the \emph{source} with $M_{\text{unl}}$ to quantify erasure.
We first run $M_{\text{full}}$ on each example to obtain reference log-probabilities $s^{\text{full}}_{i,t}$.
$M_{\text{full}}$ serves as the \emph{target} at each stage because it has learned forget set knowledge and can decode it from the patched hidden states.

\paragraph{Stage 1: Baselining.}
% To determine how much forget set knowledge is\ju{단 retain set knowledge랑 다른..} encoded at each layer\ju{each~each}, 
To establish a baseline of knowledge unique to the forget set, for each example $i$ and layer $l$, we patch the \emph{source} $M_{\text{ret}}$'s residual stream (see Appendix~\ref{app:component} for component-level analysis) into $M_{\text{full}}$ at the positions where entity tokens are predicted. We then measure the degradation in log-probability:
\begin{equation}
\Delta^{S1}_{i,l} = \frac{1}{T_i}\sum_{t=1}^{T_i}\bigl(s^{\text{full}}_{i,t} - s^{S1}_{i,t}\bigr)
\label{eq:delta-s1}
\end{equation}
% where $s^{S1}_{i,t}$ is the log-probability of entity token $y_{i,t}$ after patching layer $l$.
where $s^{S1}_{i,t}$ denotes the corresponding value for entity token $y_{i,t}$ after patching layer $l$.
A large $\Delta^{S1}_{i,l}$ indicates that $M_{\text{full}}$ encodes forget set knowledge for example $i$ at layer $l$ that $M_{\text{ret}}$ lacks.
% this pattern varies across examples, as different facts are localized at different layers

Layers with negligible $\Delta^{S1}_{i,l}$ reflect noise rather than knowledge encoding, so we set a threshold $\tau$ and keep only the \textbf{Knowledge-Encoding (KE) layers} (see Appendix~\ref{app:threshold} for sensitivity analysis):
\begin{equation}
\text{KE}_i = \{ l : \Delta^{S1}_{i,l} > \tau \}, \quad \tau = 0.05
\label{eq:ke}
\end{equation}
This also bounds the denominator $\Delta^{S1}_{i,l}$ in the Layer Erasure Ratio (Eq.~\ref{eq:ler}) away from zero.
% This also ensures that $\Delta^{S1}_{i,l}$ in the denominator of the Layer Erasure Ratio (Eq.~\ref{eq:ler}) is bounded away from zero.
% Notably, since Stage~1 depends only on $M_{\text{ret}}$ and $M_{\text{full}}$, it need only be computed once when evaluating multiple unlearned models.\ju{이 문장 implemetaion섹션으로 빼기?}

\paragraph{Stage 2: Quantification.}
We repeat the same procedure with $M_{\text{unl}}$ as the \emph{source} to quantify how much of this knowledge remains recoverable:
\begin{equation}
\Delta^{S2}_{i,l} = \frac{1}{T_i}\sum_{t=1}^{T_i}\bigl(s^{\text{full}}_{i,t} - s^{S2}_{i,t}\bigr)
\label{eq:delta-s2}
\end{equation}
If unlearning erased the knowledge for example $i$ at layer $l$, patching $M_{\text{unl}}$ should degrade predictions as much as patching $M_{\text{ret}}$: $\Delta^{S2}_{i,l} \approx \Delta^{S1}_{i,l}$.
Conversely, if the knowledge remains intact, $M_{\text{full}}$ can still decode it from the patched states, so $\Delta^{S2}_{i,l} \approx 0$.

\paragraph{Score Aggregation.}
We define the \textbf{Layer Erasure Ratio (LER)} to represent each layer's erasure as a fraction of its baseline:
\begin{equation}
\text{LER}_{i,l} = \text{clip}\!\left(\frac{\Delta^{S2}_{i,l}}{\Delta^{S1}_{i,l}},\; 0,\; 1\right)
\label{eq:ler}
\end{equation}
% where clipping to $[0,1]$ reflects that erasure to match $M_{\text{ret}}$ is the intended level.
where clipping to $[0,1]$ ensures that the metric caps at the target unlearning level defined by $M_{\text{ret}}$.

% The per-example \textsc{UDS} aggregates LER across KE layers, weighted by $\Delta^{S1}_{i,l}$ so that layers where more forget set knowledge is encoded contribute proportionally:
The per-example \textsc{UDS} aggregates LER across KE layers, weighted by $\Delta^{S1}_{i,l}$ so that layers encoding more forget set knowledge contribute proportionally:
\begin{equation}
\textsc{UDS}_i = \frac{\sum_{l \in \text{KE}_i} \Delta^{S1}_{i,l} \cdot \text{LER}_{i,l}}{\sum_{l \in \text{KE}_i} \Delta^{S1}_{i,l}}
\label{eq:uds}
\end{equation}
A score of 1 indicates knowledge erased to the level of $M_{\text{ret}}$, while 0 indicates fully intact knowledge.
If $\text{KE}_i = \emptyset$, the score is undefined and the example is excluded from aggregation.
The model-level score averages over the remaining $N$ examples:
\begin{equation}
\textsc{UDS} = \frac{1}{N}\sum_{i=1}^{N} \textsc{UDS}_i
\label{eq:uds-model}
\end{equation}

\subsection{Efficient Implementation}\label{subsec:efficient}
To efficiently evaluate large pools of unlearned models, \textsc{UDS} uses the following implementation strategy.
Because Stage~1 depends solely on $M_{\text{ret}}$ and $M_{\text{full}}$, the reference log-probabilities $s^{\text{full}}_{i,t}$, the baselines $\Delta^{S1}_{i,l}$, and the KE layer sets can be computed once and cached.
Evaluating any subsequent unlearned model requires only extracting its hidden states and running the Stage~2 patched forward passes.
Furthermore, leveraging teacher forcing across pre-identified entity spans computes all token predictions in a single forward pass per layer, avoiding the latency of autoregressive generation.
Appendix~\ref{app:runtime} reports per-model runtime, showing that these strategies keep \textsc{UDS} competitive with existing evaluation metrics.

\subsection{Validation Across Model Scales}\label{subsec:scale}
We verify that UDS reliably captures unlearning depth regardless of model scale by evaluating it across Llama 1B, 3B, and 8B \citep{grattafiori2024llama} using TOFU \citep{maini2024tofu} retain splits as source models.
% To verify that \textsc{UDS} reliably captures unlearning depth regardless of model scale, we evaluate it across Llama 1B, 3B, and 8B \citep{grattafiori2024llama} using TOFU \citep{maini2024tofu} retain splits as source models.
\begin{table}[ht]
\vspace{-0.5mm}
\centering
\small
\begin{tabular}{lrrr}
\toprule
\textbf{Source Model} & \textbf{1B} & \textbf{3B} & \textbf{8B} \\
\midrule
\texttt{full} (0\% unseen) & 0.002 & 0.008 & 0.000 \\
\texttt{retain99} (10\% unseen) & 0.153 & 0.151 & 0.101 \\
\texttt{retain95} (50\% unseen) & 0.496 & 0.482 & 0.455 \\
\texttt{retain90} (100\% unseen) & 1.000 & 1.000 & 1.000 \\
\bottomrule
\end{tabular}
\caption{\textsc{UDS} across Llama 1B, 3B, and 8B. S1 baseline is \texttt{retain90} at each scale.}
\label{tab:scale}
\vspace{-2mm}
\end{table}

As shown in Table~\ref{tab:scale}, the monotonic ordering $\texttt{full} < \texttt{retain99} < \texttt{retain95} < \texttt{retain90}$ holds at all three scales, with \textsc{UDS} values proportional to the fraction of the forget set each model has not seen.
% (\texttt{retain99}: 0.153 $\rightarrow$ 0.101, \texttt{retain95}: 0.496 $\rightarrow$ 0.455)
Values decrease slightly with scale, which is expected since larger models have greater capacity and thus a small difference in training data causes less representational shift.
For instance, removing 1\% of training data perturbs the hidden states of an 8B model less than those of a 1B model, resulting in a smaller \textsc{UDS}.
These results confirm that the monotonicity and proportionality of \textsc{UDS} remain consistent across scales.

% As shown in Table~\ref{tab:scale}, the monotonic ordering (\textit{full} $<$ \textit{retain99} $<$ \textit{retain95} $<$ \textit{retain90}) strictly holds across all three scales, with \textsc{UDS} proportionally reflecting the fraction of the unseen forget set.
% The absolute scores decrease slightly as model scale increases (e.g., \textit{retain99} drops from 0.153 to 0.101).
% This behavior is expected: larger models possess greater representational capacity, meaning the removal of a small data fraction (e.g., 1\%) induces a less severe global representational shift.
% Consequently, the baseline gap measured by activation patching slightly narrows.
% These results confirm that the monotonicity and proportionality of \textsc{UDS} remain stable regardless of model size.
\section{Meta-Evaluation}\label{sec:meta-eval}
% To validate \textsc{UDS}, we adopt the meta-evaluation framework of \citet{dorna2025openunlearning}, which provides model pools for testing\ju{provide model pool이 아니고 meta-eval이 reliable unlearning metric을 쟤는 evaluation이라는 문장이 와야함.} \emph{faithfulness}---whether a metric can distinguish models that possess forget set knowledge from those that do not---and perturbation protocols for testing \emph{robustness}---whether its scores remain stable under quantization and relearning.
% \ju{아래 문장 제거}
% \S\ref{subsec:meta-setup} describes the models and dataset, comparison metrics, and the evaluation criteria including our symmetric robustness formulas, and \S\ref{subsec:results} analyzes the results.
To validate \textsc{UDS}, we adopt and extend the meta-evaluation framework of \citet{dorna2025openunlearning}, which tests metric faithfulness and robustness.

\subsection{Setup}\label{subsec:meta-setup}
% \paragraph{Models and Dataset.}
% \ju{문단 삭제}
% \ju{아키텍쳐, 벤치마크 데이터셋->intro, unlearned model/method/sweep -> robustness, Mretain/full->?}
% We evaluate on the TOFU forget10 benchmark \citep{maini2024tofu} using Llama-3.2-1B-Instruct \citep{grattafiori2024llama} as the base architecture.
% Our evaluation spans 150 unlearned models produced by 8 methods \ju{(2.1에서 언급한 메소드 on the full model? Mfull먼저?)} across hyperparameter sweeps (see Appendix~\ref{app:unlearning} for details).
\paragraph{Models and Dataset.}
We evaluate on the TOFU forget10 benchmark \citep{maini2024tofu} using the Llama-3.2-1B-Instruct \citep{grattafiori2024llama} architecture. The faithfulness evaluation uses a P-pool (30 models trained on data including $D_f$) and an N-pool (30 models trained without $D_f$). The robustness evaluation spans 150 unlearned models produced by 8 methods described in \S\ref{sec:related} across hyperparameter sweeps (see Appendix~\ref{app:unlearning} for details). $M_{\text{full}}$ and $M_{\text{ret}}$ serve as reference models.

% \paragraph{Models and Dataset.}
% We evaluate on the TOFU forget10 benchmark \citep{maini2024tofu} using the Llama-3.2-1B-Instruct \citep{grattafiori2024llama} architecture. 
% The faithfulness evaluation uses a P-pool (30 models trained on data including $D_f$) and an N-pool (30 models trained without $D_f$). The robustness evaluation spans 150 unlearned models produced by 8 methods across hyperparameter sweeps: GradDiff \citep{yao2024large}, NPO \citep{zhang2024negative}, SimNPO \citep{fan2025simplicity}, AltPO \citep{mekala2025alternate}, IdkNLL, IdkDPO \citep{maini2024tofu}, RMU \citep{li2024wmdp}, and UNDIAL \citep{dong2025undial} (see Appendix~\ref{app:unlearning} for details). 
% $M_{\text{full}}$ and $M_{\text{ret}}$ serve as reference models.

\paragraph{Comparison Metrics.}
We compare \textsc{UDS} against 19 metrics. 
Twelve are from \citet{dorna2025openunlearning}: eight memorization metrics (ES, EM, Prob, ParaProb, Truth Ratio, ROUGE, Para-ROUGE, Jailbreak-ROUGE) and four MIA variants (LOSS, ZLib, Min-K, Min-K++).

Since \textsc{UDS} is retain-referenced and operates on internal representations, we add four retain-referenced MIA variants and three white-box baselines.
% Since \textsc{UDS} is both retain-referenced and internal \jj{문장 어색 internal?}, we add four retain-referenced MIA variants and three white-box baselines for fair comparison.
The retain-referenced MIA variants ($s_{\text{LOSS}}$, $s_{\text{ZLib}}$, $s_{\text{Min-K}}$, $s_{\text{Min-K++}}$) scale raw MIA AUC against $M_{\text{ret}}$, adapting PrivLeak normalization \citep{shi2025muse}:
\begin{equation}
s_* = 1 - \min\!\left(\frac{|\text{AUC}_m - \text{AUC}_{\text{ret}}|}{\text{AUC}_{\text{ret}}},\; 1\right)
\label{eq:smia}
\end{equation}
The white-box baselines compare $M_{\text{unl}}$ and $M_{\text{ret}}$ at each layer, aggregated as $\sum_l w_l \, e_l \,/\, \sum_l w_l$ where $w_l$ captures layer importance and $e_l$ measures erasure:
CKA \citep{kornblith2019similarity} (representational similarity), Logit Lens \citep{nostalgebraist2020logitlens} (frozen-decoder readout), and Fisher \citep{kirkpatrick2017overcoming} (parameter sensitivity; top $0.1\%$ mask).
Formal definitions appear in Appendix~\ref{app:metrics}.
\paragraph{Evaluation Protocol.}
To assess faithfulness, we compute the AUC-ROC to measure how well each metric separates the P-pool from the N-pool.
\begin{figure}[t]
    \centering
    \includegraphics[width=\columnwidth]{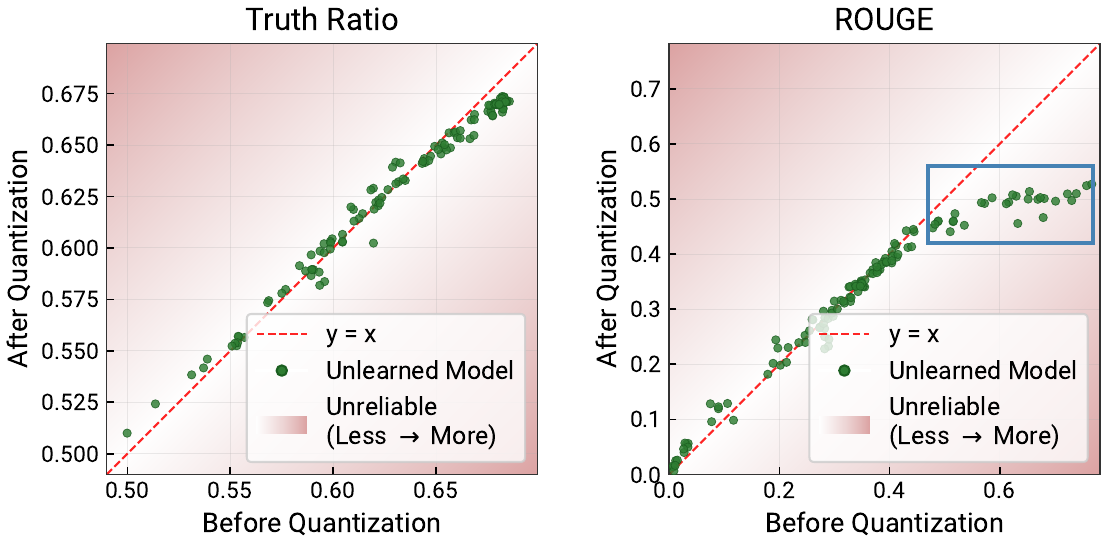}
\caption{Quantization test for Truth Ratio and ROUGE.
Truth Ratio dots lie along the diagonal, indicating stable scores; ROUGE dots in the blue box fall below the diagonal.
One-directional formulas do not penalize this decline, but our symmetric formula (Eq.~\ref{eq:qr}) does.}
    \label{fig:robustness}
\end{figure}
For robustness, we evaluate metric stability under 4-bit quantization and 1-epoch relearning on $D_f$. 
% (see Appendix~\ref{app:attack-settings} for details).
\citet{dorna2025openunlearning} score quantization stability as $\min(m/m',1)$ and relearning stability as $\min(\Delta_{\text{ret}}/\Delta_{\text{unl}},1)$, 
% where $m$ ($m'$) is the metric value before (after) purturbation and $\Delta = m' - m$.
where $m$ ($m'$) is the metric value before (after) the intervention, oriented so that higher values indicate knowledge retention, and $\Delta = m' - m$.
% These formulations effectively penalize knowledge recovery but not score decreases.
% For instance, quantization can cause ROUGE to decline regardless of knowledge content (Figure~\ref{fig:robustness}, blue box), yet such decline is rewarded by construction.
% We therefore propose symmetric alternatives to penalize changes in both directions (i.e., score increases and decreases):
These asymmetric formulations penalize knowledge recovery (i.e., score increases) but overlook spurious metric degradation (e.g., score drops caused by impaired generation).
For instance, quantization can degrade generation quality by reducing model precision, causing ROUGE to decline regardless of knowledge content (Figure~\ref{fig:robustness}, blue box), yet such decline is rewarded by construction.
We therefore propose symmetric alternatives to penalize changes in both directions:
\begin{equation}
Q = 1 - \frac{|m' - m|}{|m'| + |m|}, R = 1 - \frac{|\Delta_{\text{unl}} - \Delta_{\text{ret}}|}{|\Delta_{\text{unl}}| + |\Delta_{\text{ret}}|}
\label{eq:qr}
\end{equation}
In practice, a small constant is added to the denominators to prevent division by zero.
For the normalized MIA scores and the four white-box metrics, where higher values indicate erasure, we apply $m \leftarrow 1{-}m$ prior to computation.
\begin{figure}[t]
    \centering
    \includegraphics[width=\columnwidth]{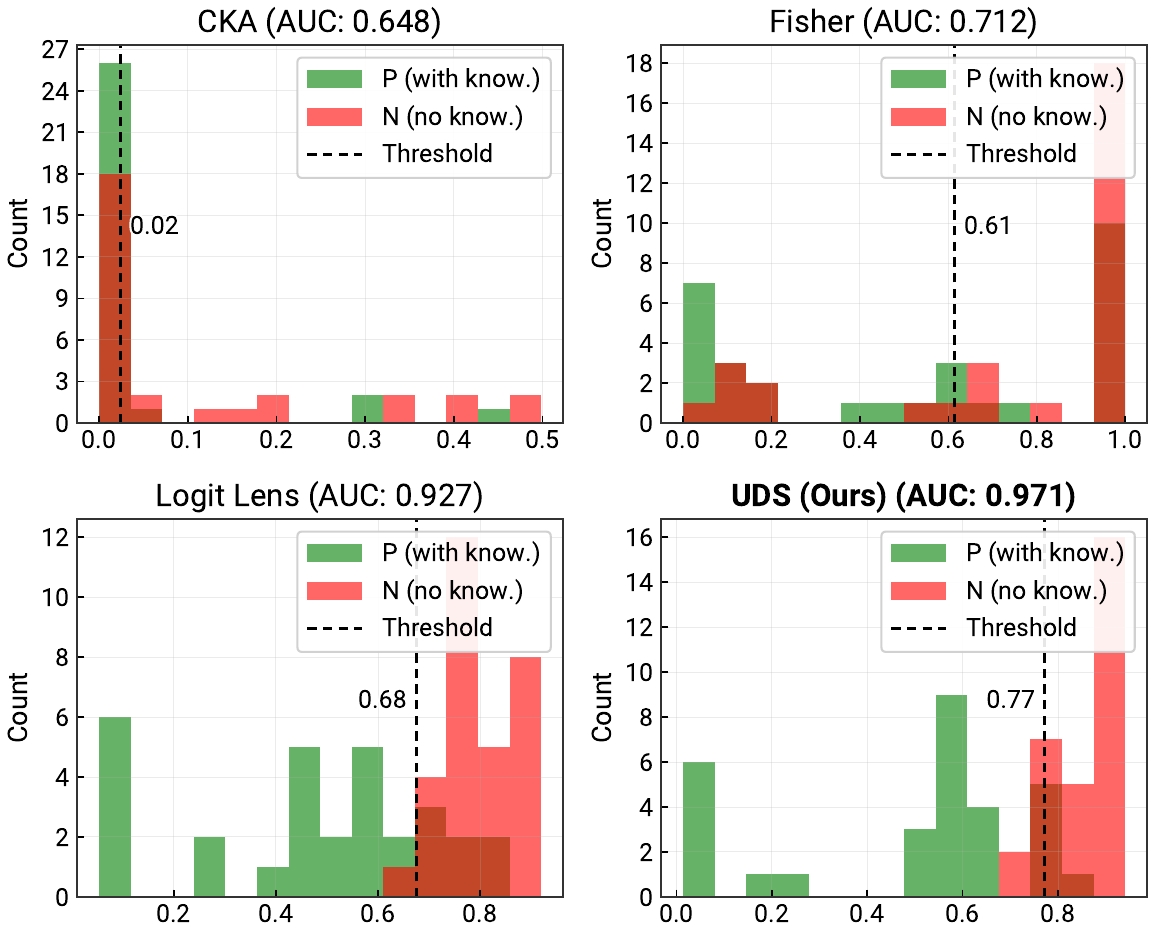}
\caption{Faithfulness evaluation of four white-box metrics via P/N pool separation.
CKA measures geometry and Fisher measures gradient sensitivity, neither of which directly reflects knowledge content, yielding poor separation.
Logit Lens and \textsc{UDS} distinguish the pools well; \textsc{UDS} further benefits from causal intervention.}
    \label{fig:faithfulness}
% \vspace{-2mm}
\end{figure}

% Requires: \usepackage{booktabs, xcolor, pifont, array}
% Preamble commands:
% \newcommand{\best}[1]{\textbf{#1}}
% \newcommand{\second}[1]{\underline{#1}}
% \newcommand{\worst}[1]{\textbf{\textcolor{red!70!black}{#1}}}
% \newcommand{\secworst}[1]{\textcolor{red!70!black}{\underline{#1}}}
% \newcommand{\cmark}{\ding{51}}

\begin{table*}[t]
\centering
\small
\setlength{\tabcolsep}{4pt}
\begin{tabular}{@{}ll ccc w{r}{3.6em}w{r}{3.6em}w{r}{3.6em}w{r}{3.6em}w{r}{3.6em}@{}}
\toprule
 & & \multicolumn{3}{c}{\textbf{Scope}} & & & \multicolumn{3}{c}{\textbf{Robustness}} \\
\cmidrule(lr){3-5} \cmidrule(l){8-10}
\textbf{Group} & \textbf{Metric}
  & \textbf{O} & \textbf{R} & \textbf{I}
  & \textbf{Overall\,$\uparrow$} & \textbf{Faith.\,$\uparrow$}
  & \textbf{Agg.\,$\uparrow$} & \textbf{Quant.\,$\uparrow$} & \textbf{Relearn\,$\uparrow$} \\
\midrule
\textbf{Memorization}
  & Extraction Strength & \cmark & & & 0.875 & 0.891 & 0.859 & 0.970 & 0.770 \\
  & Exact Memorization  & \cmark & & & 0.782 & 0.817 & 0.750 & 0.984 & 0.605 \\
  & Probability         & \cmark & & & 0.786 & 0.816 & 0.757 & 0.924 & 0.642 \\
  & Paraphrased Probability & \cmark & & & 0.782 & \secworst{0.707} & 0.875 & 0.853 & 0.899 \\
  & Truth Ratio         & \cmark & & & 0.542 & \second{0.947} & 0.379 & \second{0.996} & 0.234 \\
\cmidrule(l){2-10}
\textit{\quad Generation}
  & ROUGE               & \cmark & & & 0.456 & 0.722 & 0.333 & 0.934 & 0.203 \\
  & Paraphrased ROUGE    & \cmark & & & \secworst{0.209} & 0.832 & \secworst{0.119} & 0.951 & \secworst{0.064} \\
  & Jailbreak ROUGE     & \cmark & & & 0.438 & 0.757 & 0.308 & 0.971 & 0.183 \\
\midrule
\textbf{Privacy}
  & MIA-LOSS            & \cmark & & & 0.767 & 0.902 & 0.668 & 0.935 & 0.519 \\
  & MIA-ZLib            & \cmark & & & 0.737 & 0.867 & 0.641 & 0.938 & 0.487 \\
  & MIA-Min-K           & \cmark & & & 0.774 & 0.907 & 0.675 & 0.923 & 0.532 \\
  & MIA-Min-K++         & \cmark & & & 0.677 & 0.816 & 0.579 & 0.883 & 0.431 \\
\cmidrule(l){2-10}
\textit{\quad Normalized}
  & $s_{\text{LOSS}}$     & \cmark & \cmark & & 0.778 & 0.891 & 0.690 & 0.719 & 0.663 \\
  & $s_{\text{ZLib}}$     & \cmark & \cmark & & 0.790 & 0.870 & 0.724 & 0.704 & 0.745 \\
  & $s_{\text{Min-K}}$    & \cmark & \cmark & & 0.786 & 0.891 & 0.704 & 0.710 & 0.697 \\
  & $s_{\text{Min-K++}}$  & \cmark & \cmark & & 0.686 & 0.799 & 0.602 & \secworst{0.643} & 0.566 \\
\cmidrule(l){2-10}
\textit{\quad White-box}
  & CKA                   & & \cmark & \cmark & \worst{0.051} & \worst{0.648} & \worst{0.026} & \best{0.997} & \worst{0.013} \\
  & Fisher (Masked 0.1\%) & & \cmark & \cmark & 0.716 & 0.712 & 0.721 & \worst{0.583} & \best{0.946} \\
  & Logit Lens            & \cmark & \cmark & \cmark & \second{0.902} & 0.927 & \second{0.879} & 0.959 & 0.812 \\
  & \textbf{\textsc{UDS} (Ours)} & \cmark & \cmark & \cmark & \best{0.951} & \best{0.971} & \best{0.932} & 0.968 & \second{0.900} \\
\bottomrule
\end{tabular}
\caption{Meta-evaluation of 20 unlearning metrics, with \best{best}, \second{second}, \worst{worst}, and \secworst{second-worst} marked in each column.
Scope tags mark the signals a metric uses: output-level (\textbf{O}), retain-referenced (\textbf{R}), or internal (\textbf{I}).
Each metric is scored on faithfulness (AUC-ROC) and robustness (harmonic mean of quantization and relearning); Overall combines both via harmonic mean.
\textsc{UDS} ranks first in Overall, Faithfulness, and Aggregate Robustness.}
\label{tab:meta-eval}
\end{table*}

\subsection{Results}\label{subsec:results}
Following \citet{dorna2025openunlearning}, we restrict robustness evaluation to models that preserve at least 80\% utility of $M_{\text{full}}$ and are classified as unlearned by the metric's faithfulness threshold.
Per-metric robustness is the harmonic mean (HM) of $Q$ and $R$ averaged across the models, and the overall score combines faithfulness and robustness via harmonic mean. Table~\ref{tab:meta-eval} reports full results for all 20 metrics; per-metric plots appear in Appendix~\ref{app:full-plots}.

\paragraph{Faithfulness.}
\textsc{UDS} achieves the highest faithfulness (AUC 0.971).
The second best and top output-level metric is Truth Ratio (0.947), in line with \citet{dorna2025openunlearning}.
Notably, the white-box baselines diverge.
% CKA (0.648) and Fisher (0.712) achieve poor separation: CKA measures how similarly two models represent the same dataset, but unlearning can alter representational geometry without removing specific knowledge, so low similarity does not entail erasure; Fisher measures gradient sensitivity, which has been shown to reflect optimization trajectories rather than true knowledge content \citep{basu2021influence}.
CKA (0.648) and Fisher (0.712) achieve poor separation.
CKA measures how similarly two models represent the same dataset, but unlearning can alter representational geometry without removing knowledge, so higher similarity to $M_\text{ret}$ does not mean erasure.
Fisher measures gradient sensitivity, which reflects optimization trajectories rather than knowledge content \citep{basu2021influence}.
Logit Lens (0.927) reads knowledge through the frozen decoder, achieving strong separation; \textsc{UDS} improves upon this through causal intervention (Figure~\ref{fig:faithfulness}).
The AUC-optimal boundary between the two pools is $\textsc{UDS} = 0.77$, which practitioners can adopt as a reference threshold for classifying a model as erased.

\paragraph{Robustness.}
\textsc{UDS} ranks first in aggregate robustness ($\operatorname{HM}=0.932$), with balanced quantization ($Q=0.968$) and relearning ($R=0.900$) stability.
Unlike output metrics that are easily disrupted by output-level distribution shifts, \textsc{UDS} extracts intermediate representations and evaluates them through the unperturbed computational pathways of $M_{\text{full}}$.
By bypassing the unlearned model's output head, \textsc{UDS} is resilient to weight compression.
Furthermore, it remains stable under relearning: while a single epoch of fine-tuning triggers rapid generation recovery by realigning the output space, it does not substantially alter the deeply encoded knowledge that \textsc{UDS} measures.
Logit Lens follows ($\operatorname{HM}=0.879$), though its fixed-decoder readout is less stable under relearning ($R=0.812$).

By contrast, CKA collapses under relearning ($R=0.013$) because global representational geometry shifts even under brief fine-tuning.
Fisher is the most vulnerable to quantization ($Q=0.583$), below even the MIA variants ($Q{\geq}0.643$, the least quantization-stable among output-level metrics), because weight compression distorts the gradient landscape it relies on.
Among output-level metrics, ROUGE variants are highly unstable under relearning ($R=0.06$--$0.20$): residual knowledge in the unlearned model enables rapid recovery of generation, far exceeding what the retain model learns for the first time in one epoch.
Truth Ratio, despite the second-highest faithfulness, degrades sharply under relearning ($R=0.234$).

Overall, \textsc{UDS} achieves the highest score (0.951) across all 20 metrics, confirming that the causal approach provides the most reliable evaluation of knowledge erasure.
\section{Case Studies}\label{sec:case-study}
\textsc{UDS} provides per-layer, per-example erasure scores, enabling analyses beyond aggregate evaluation.
We present two analyses: observational vs.\ causal evaluation (\S\ref{subsec:obs-causal}), and prompt-type variation in erasure depth within a single model (\S\ref{subsec:heterogeneity}).

\subsection{Observational vs.\ Causal Evaluation}\label{subsec:obs-causal}
% Requires: \usepackage{booktabs, array}

\begin{table}[t]
\centering
\small
\setlength{\tabcolsep}{4pt}
\begin{tabular}{@{}c rrr rrr@{}}
\toprule
& \multicolumn{3}{c}{\textbf{Logit Lens}} & \multicolumn{3}{c}{\textbf{UDS (Ours)}} \\
\cmidrule(lr){2-4} \cmidrule(l){5-7}
\textbf{Layer} & $\Delta^{S1}$ & $\Delta^{S2}$ & LER
               & $\Delta^{S1}$ & $\Delta^{S2}$ & LER \\
\midrule
0--4 & \multicolumn{3}{c}{\textit{not KE}} & \multicolumn{3}{c}{\textit{not KE}} \\
5   & 0.375 & 0.250 & 0.667 & \multicolumn{3}{c}{\textit{not KE}} \\
7   & 0.312 & 0.812 & \textbf{1.000} & 0.053 & $-$0.059 & \textbf{0.000} \\
9   & 1.375 & 1.375 & \textbf{1.000} & 0.346 & 0.039 & \textbf{0.113} \\
11  & 1.250 & 2.062 & \textbf{1.000} & 0.838 & 0.088 & \textbf{0.105} \\
13  & 0.926 & 2.465 & \textbf{1.000} & 1.299 & 0.299 & \textbf{0.230} \\
15  & 1.713 & 0.436 & 0.254 & 1.713 & 0.436 & 0.254 \\
\midrule
\multicolumn{1}{@{}l}{\textbf{Score}}
    & & & \textbf{0.801} & & & \textbf{0.209} \\
\bottomrule
\end{tabular}
\caption{Layer-wise comparison on a forget set example from an IdkDPO model.
Logit Lens judges the knowledge as erased (0.801)
but \textsc{UDS} does not (0.209); observational decoding misses knowledge
that causal intervention identifies as recoverable.}
\label{tab:case-ll-uds}
\end{table}

As analyzed in \S\ref{subsec:results}, the causal approach of \textsc{UDS} achieves higher faithfulness than the observational readout of Logit Lens.
We examine a specific forget set example to understand the mechanism driving this discrepancy.
Because Logit Lens relies on a frozen decoder, it is vulnerable to representational shifts: if an unlearning method rotates or distorts the internal vector space, the fixed unembedding matrix fails to read the forget set knowledge, and the metric falsely concludes it has been erased.
\textsc{UDS} overcomes this limitation because its causal patching allows the remaining nonlinear layers of $M_{\text{full}}$ to process and realign these distorted vectors.

Table~\ref{tab:case-ll-uds} illustrates this gap on an IdkDPO model predicting the entity ``historical fiction.''
Logit Lens reports a high aggregate erasure score of 0.801, indicating complete erasure ($\text{LER}=1.000$) from layers 7 through 13.
In contrast, \textsc{UDS} yields an overall score of 0.209, revealing that the forget set knowledge remains highly recoverable ($\text{LER} \le 0.230$) across these mid-layers.
Both metrics converge to an identical Layer Erasure Ratio (0.254) at the final layer, where representations directly determine the output logits.
This layer-wise divergence demonstrates that while representational shifts easily mislead observational metrics, \textsc{UDS} confirms the underlying knowledge remains intact.

\subsection{Heterogeneity of Unlearning Depth}\label{subsec:heterogeneity}

Even within a single unlearning method, unlearning depth varies substantially across examples.
We illustrate this with IdkNLL, which unlearns by replacing correct answers with ``I don't know.''
All configurations score approximately 0.0 on every normalized MIA variant, yet \textsc{UDS} differentiates them, ranging from 0.039 to 0.253.
Table~\ref{tab:prompt-type} reports per-prompt-type scores for one model: Yes/No questions reach 0.624, while all other types (e.g., person names, book titles) fall below 0.05.
% For Yes/No questions, ``I don't know'' is semantically related to negating the answer, so the substitution modifies knowledge-encoding layers.
For Yes/No questions, ``I don't know'' functions as a negation of the factual answer, modifying deeper KE layers.
For other prompt types, this response bears no semantic relation to the original entity, and only the output distribution changes.
Per-example \textsc{UDS} enables identifying which knowledge types a given method fails to erase internally.

% Requires: \usepackage{booktabs, makecell}

\begin{table}[t]
\centering
\small
\setlength{\tabcolsep}{3pt}
\begin{tabular}{@{}l r l @{\hskip -4pt} r@{}}
\toprule
\textbf{Prompt Type} & \textbf{N} & \textbf{Example Entity}
  & \textbf{Mean UDS} \\
\midrule
\textbf{Yes/No} & 21 & Yes
  & \textbf{0.624} \\
Person Name & 15 & Hsiao Yun-Hwa
  & 0.025 \\
Book/Title & 88
  & \makecell[l]{``Artistic Authority:\\[-2pt]Leading with Creativity''}
  & 0.038 \\
Biographical & 75
  & \makecell[l]{dietician; 2002; Seoul\\[-2pt]Leadership Literary Award}
  & 0.049 \\
Descriptive & 162
  & \makecell[l]{cultural understanding,\\[-2pt]inclusivity and diversity}
  & 0.042 \\
\midrule
\textbf{Overall} & \textbf{361} & ---
  & \textbf{0.076} \\
\bottomrule
\end{tabular}
\caption{Per-prompt-type \textsc{UDS} from a single IdkNLL model.
Yes/No questions show substantially higher erasure ($0.624$) than all other types ($0.025$--$0.049$), revealing that erasure depth varies by prompt type.}
\label{tab:prompt-type}
\end{table}

\section{Practical Implications}\label{sec:implications}
We discuss three practical implications of \textsc{UDS}: integrating it into the privacy axis (\S\ref{subsec:privacy-axis}), streamlining the evaluation pipeline (\S\ref{subsec:pipeline}), and applying \textsc{UDS} when no retain model is available
(\S\ref{subsec:retain-free}).

\subsection{Integrating \textsc{UDS} into the Privacy Axis}\label{subsec:privacy-axis}
% \jj{제목은 좋은데 실제로 본문은 reranking과 튜닝에 집중하고 있음. 제목이 모두를 포괄하지 못하는 느낌}
Existing frameworks evaluate the privacy axis primarily through output-level MIA metrics \citep{shi2025muse, jin2024rwku, dorna2025openunlearning}; for instance, MUSE defines $\text{Privacy} = s_{\text{Min-K}}$.
However, the risk of latent knowledge extraction \citep{lynch2024eight} highlights the need for a broader definition.
We suggest extending this by defining:
\begin{equation}
\text{Privacy} = \text{HM}(\text{MIA}_{\text{agg}},\; \textsc{UDS}) \label{eq:privacy}
% \text{MIA}_{\text{agg}} &= \text{HM}(s_{\text{LOSS}},\, s_{\text{ZLib}},\, s_{\text{Min-K}},\, s_{\text{Min-K++}})   
\end{equation}
where $\text{MIA}_{\text{agg}}$ is the harmonic mean of four MIA variants from \citet{dorna2025openunlearning}, each normalized against $M_{\text{ret}}$ (Eq.~\ref{eq:smia}).
This harmonic formulation penalizes one-sided degradation, encouraging deeper internal erasure while maintaining output-level privacy.

\paragraph{Impact on Method Ranking.}
% Requires: \usepackage{booktabs}

\begin{table}[t]
\centering
\small
\setlength{\tabcolsep}{5.5pt}
\begin{tabular}{@{}l @{\hskip 8pt} r r r r@{}}
\toprule
\textbf{Method} & \textbf{w/o (Rank)} & \textbf{w/ (Rank)} & $\textbf{MIA}_{\textbf{agg}}$ & \textbf{\textsc{UDS}} \\
\midrule
AltPO    & 0.784\,(1)                      & 0.766\,(1) & 0.952 & 0.816 \\
\textbf{SimNPO}   & 0.733\,(3) & 0.722\,(\textbf{3$\to$2}) & 0.816 & \textbf{0.739} \\
\textbf{NPO}      & 0.752\,(2) & 0.710\,(\textbf{2$\to$3}) & 0.875 & \textbf{0.619} \\
IdkDPO   & 0.720\,(4)                      & 0.709\,(4) & 0.757 & 0.686 \\
GradDiff & 0.686\,(5)                      & 0.637\,(5) & 0.789 & 0.515 \\
RMU      & 0.631\,(6)                      & 0.625\,(6) & 0.711 & 0.667 \\
UNDIAL   & 0.088\,(7)                      & 0.103\,(7) & 0.091 & 0.871 \\
IdkNLL   & 0.000\,(8)                      & 0.000\,(8) & 0.000 & 0.251 \\
\bottomrule
\end{tabular}
\caption{
Method ranking before and after integrating \textsc{UDS} into the privacy axis.
Each score is HM(Memorization, Privacy, Utility); see Appendix~\ref{app:aggregation} for axis definitions.
Best configuration per method selected by the \textbf{w/o} formula.
\textbf{w/o}: Privacy $=$ $\text{MIA}_{\text{agg}}$;
\textbf{w/}: Privacy $=$ HM($\text{MIA}_{\text{agg}}$, \textsc{UDS}).
Adding \textsc{UDS} swaps NPO and SimNPO, exposing internal erasure differences that $\text{MIA}_{\text{agg}}$ alone misses.}

\label{tab:ranking}
\end{table}

Table~\ref{tab:ranking} compares method rankings before and after integrating \textsc{UDS} into the privacy axis.
% \jj{위에서 하나로 합쳤는데, the two라고 합치기전을 얘기하니까 어색함}
Adding \textsc{UDS} to the privacy axis causes NPO and SimNPO to swap ranks (2$\leftrightarrow$3).
NPO's best configuration, selected without \textsc{UDS}, achieves a high $\text{MIA}_{\text{agg}}$ score (0.875) but moderate internal erasure (\textsc{UDS} = 0.619).
Conversely, the SimNPO configuration yields a lower $\text{MIA}_{\text{agg}}$ score but a higher \textsc{UDS} (0.739); its length-normalized, reference-free objective drives unlearning pressure deeper into intermediate representations.
While establishing fundamental algorithmic superiority requires broader investigation, this configuration-level rank swap illustrates how \textsc{UDS} complements existing evaluations by exposing residual internal knowledge that output-only metrics overlook.

\paragraph{Impact on Hyperparameter Selection.}
% Requires: \usepackage{booktabs}

\begin{table}[t]
\centering
\small
\setlength{\tabcolsep}{3.5pt}
\begin{tabular}{@{}l ccc ccc r@{}}
\toprule
& \multicolumn{3}{c}{\textbf{Best by w/o \textsc{UDS}}} & \multicolumn{3}{c}{\textbf{Best by w/ \textsc{UDS}}} & \\
\cmidrule(lr){2-4} \cmidrule(lr){5-7}
\textbf{Method} & LR & $\alpha$ & Epoch & LR & $\alpha$ & Epoch & \textbf{$\Delta$\textsc{UDS}} \\
\midrule
AltPO & 5e-5 & 1 & 5  & 5e-5 & 2 & \textbf{10}  & \textbf{+0.016} \\
NPO   & 2e-5 & 1 & 10 & \textbf{5e-5} & 5 & 10  & \textbf{+0.199} \\
\bottomrule
\end{tabular}
\caption{Two configuration shifts under the \textbf{w/} \textsc{UDS} formula.
$\alpha$ is the retention coefficient.
Both methods shift toward higher learning rates or longer unlearning, improving \textsc{UDS}.}
\label{tab:config-shift}
\end{table}

\textsc{UDS} also reshapes which configurations practitioners would select.
Table~\ref{tab:config-shift} shows two examples: when re-selecting the best configuration with the w/~\textsc{UDS} formula, both AltPO and NPO shift toward higher learning rates or longer training.
These shifts demonstrate that integrating \textsc{UDS} steers hyperparameter selection toward deeper internal erasure.
% \jj{이거하면 부작용이 없는지? 즉 MIA-agg같은 것은 거의 유지가 되는지? 단순 트레이드오프를 보이는 것 아닌지?}
% \textsc{UDS} also reshapes which practitioners would select.
% Table~\ref{tab:config-shift} shows two examples: when re-selecting the best configuration under the \textbf{w/} formula, both AltPO and NPO shift toward higher learning rates or longer training.
% For NPO, the shift improves internal erasure substantially ($\Delta$\textsc{UDS} = +0.199).

 % \jj{of what?}
\subsection{Streamlining the Evaluation Pipeline}\label{subsec:pipeline}
Under current evaluation frameworks, verifying the robustness of unlearning typically requires applying perturbations such as quantization and relearning to each model before re-running the evaluation suite.
Repeating this process across a large pool of models creates substantial computational overhead.
In our meta-evaluation (\S\ref{sec:meta-eval}), \textsc{UDS} demonstrates high stability under both quantization and relearning, which drives its highest aggregate robustness.
% Because of this stability, a pre-perturbation \textsc{UDS} score serves as a strong predictor of robust knowledge erasure under deployment perturbations, allowing practitioners to skip exhaustive post-perturbation benchmarking and streamline the overall evaluation pipeline.
This stability makes the pre-perturbation \textsc{UDS} score a strong predictor of robust knowledge erasure under deployment perturbations, allowing practitioners to skip exhaustive post-perturbation benchmarking and streamline the overall evaluation pipeline.

\subsection{Evaluating without a Retain Model}
\label{subsec:retain-free}
\begin{figure}[t]
\centering
\includegraphics[width=\columnwidth]{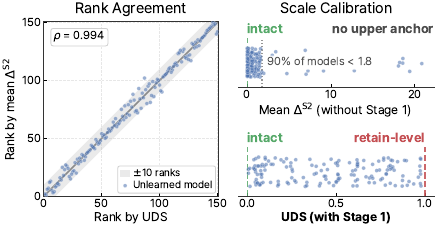}
\caption{Evaluating \textsc{UDS} without a retain model on 150 unlearned models.
Ranking by the mean $\Delta^{\mathrm{S2}}$ tracks ranking by \textsc{UDS}, with 96.7\% inside the shaded $\pm 10$-rank band (left). 
Without Stage 1 the score has no upper anchor (upper right), whereas \textsc{UDS} spans $[0, 1]$ between intact and retain-level erasure (lower right), indicating that Stage 1 calibrates the scale, not the ordering.}

\label{fig:retain-free}
\end{figure}

% \textsc{UDS} is defined against a retain model reference, but such a model is often unavailable in production.
% Because $M_\text{ret}$ enters only in Stage 1, Stage 2 can run alone, patching $M_\text{unl}$'s hidden states into $M_\text{full}$.
% Figure~\ref{fig:retain-free} compares this fallback against full \textsc{UDS} on 150 unlearned models.
% Rankings are nearly unchanged (Spearman $\rho = 0.994$); what Stage 1 adds is the interpretable scale, placing scores on $[0, 1]$ where 1 denotes erasure to the level of $M_\text{ret}$. 
% With that anchor, a single score can be read as an absolute level of erasure, not only against other models.
% When no retain model is available, Stage 2 alone therefore remains a sound basis for comparing methods and selecting among configurations.
\textsc{UDS} is defined relative to a retain model, but such a model is often unavailable in production. 
$M_\text{ret}$ enters only in Stage 1, so Stage 2 can run alone, patching $M_\text{unl}$'s hidden states into $M_\text{full}$ and averaging $\Delta^{S2}$ over all layers and examples.

Figure~\ref{fig:retain-free} compares this fallback with two-stage \textsc{UDS} on 150 unlearned models. 
Rankings are nearly unchanged (Spearman $\rho = 0.994$); what Stage 1 adds is the interpretable scale, placing scores on $[0, 1]$ anchored at retain-level erasure.
That anchor makes a single score meaningful on its own. 
Without a retain model, Stage 2 alone therefore remains a sound basis for comparing methods and selecting among configurations.

\section{Conclusion}\label{sec:conclusion}
We present \textsc{Unlearning Depth Score} (\textsc{UDS}), a metric that quantifies the mechanistic depth of unlearning by measuring knowledge recoverability through two-stage activation patching.
\textsc{UDS} produces 0--1 erasure scores, complementing output-level metrics with causal intervention.
In our meta-evaluation of 20 metrics on 150 unlearned models, \textsc{UDS} achieves the highest faithfulness and robustness, demonstrating both reliable knowledge detection and stability under deployment perturbations.
% Our case studies show that \textsc{UDS} captures knowledge retention invisible to output-level metrics, and its robustness makes it a practical substitute for costly post-perturbation evaluation pipelines.
Our case studies show that \textsc{UDS} uncovers residual knowledge invisible to output-level metrics, and its robustness makes it a practical substitute for costly post-perturbation evaluation pipelines.
% We hope \textsc{UDS} contributes to more rigorous evaluation standards for machine unlearning research. \jj{이 문장은 보통 예의상 넣는건지? 불필요해 보임. 앞에서 달려온 스타일과도 안맞고 뒤에 내용이 더있어서 어색함}
\section*{Limitations}

% \paragraph{Scope and Dataset.}
% \textsc{UDS} requires access to a gold-standard retain model.
% This assumption may not hold in all deployment contexts, but \textsc{UDS} remains useful for method development, internal audits, and benchmark curation.
% Our current evaluation focuses on the TOFU \citep{maini2024tofu}; validating on other unlearning benchmarks \citep[e.g.,][]{shi2025muse, li2024wmdp} would further strengthen generality.

% \paragraph{Requirement of a Retain Model.}
% For rigorous quantification, \textsc{UDS} employs two-stage patching, which requires a retain model.
% While \textsc{UDS} remains useful for method development, internal audits, and benchmark curation, such a model may not be available in all deployment contexts.
% Without a retain model, practitioners can utilize the quantification stage alone, patching hidden states from the unlearned model into the full model (i.e., the original model prior to unlearning).
% Although this omits retain-based normalization, observing the degradation in prediction probability still serves as a causal indicator of residual knowledge.
\paragraph{Requirement of a Retain Model.}
For rigorous quantification, \textsc{UDS} employs two-stage patching, which requires a retain model.
While \textsc{UDS} remains useful for method development, internal audits, and benchmark curation, such a model may not be available in all deployment contexts.
Without a retain model, practitioners can run Stage 2 alone; \S\ref{subsec:retain-free} shows that this preserves model rankings, though without the calibrated $[0, 1]$ scale.

% Without a retain model, practitioners can adapt our framework by utilizing the quantification stage alone, patching hidden states from the unlearned model into the full model (i.e., the original model prior to unlearning).
% Although this omits retain-based normalization, observing the degradation in prediction probability still serves as a causal indicator of residual knowledge.

\paragraph{Clipping and Over-Unlearning.}
The $\text{clip}(\cdot, 0, 1)$ operation in Eq.~\ref{eq:ler} caps \textsc{UDS} at 1.0, so over-unlearning (representations deviating beyond the retain model) is mathematically indistinguishable from perfect unlearning.
Practitioners should jointly monitor \textsc{UDS} with the utility axis (i.e., the preservation of general capabilities) to diagnose such cases.

\paragraph{Dataset.}
To maintain methodological alignment with the meta-evaluation framework of \citet{dorna2025openunlearning}, our current evaluation focuses on the TOFU benchmark \citep{maini2024tofu}.
Validating \textsc{UDS} on other unlearning benchmarks \citep[e.g.,][]{shi2025muse, li2024wmdp} would further strengthen its generality across diverse domains.

\paragraph{Entity Span.}
\textsc{UDS} currently operates on localized entity spans under teacher forcing.
How it extends to long-form or multi-step reasoning targets beyond factoid entities remains an open question.
Our automatic extraction pipeline (Appendix~\ref{app:dataset}) handles structured QA pairs but may not generalize to all data formats.

% \jj{이거 인용 없으면 무슨 말인지 모를듯}
% \paragraph{Computational Cost.} \jj{이건 너무나 당연한 얘기고 이론적으로 극복이 불가능해서 불필요해보임. 리뷰어가 지적할만한 justfication이나 quality에 대한 사전 방어로 대체하는게 나을듯}
% With the efficient implementation in \S\ref{subsec:efficient}, \textsc{UDS} requires $L{+}1$ forward passes per example (one for source model hidden state extraction plus $L$ for layer-wise patching).
% While this is faster than generation-based metrics such as ROUGE, it is considerably slower than single-pass output metrics (e.g., token log-probabilities).

\section*{Broader Impact}

\textsc{UDS} enables the detection of incomplete unlearning, supporting responsible AI deployment and regulatory compliance efforts.
However, to avoid a false sense of security, \textsc{UDS} should be deployed as part of a comprehensive safety pipeline rather than viewed as an absolute guarantee of complete data removal.

The two-stage activation patching framework underlying \textsc{UDS} is not specific to autoregressive language models.
Any architecture with layered representations (e.g., diffusion models, vision transformers) could in principle be audited for residual knowledge through analogous patching procedures.
Adapting the concrete metric formulation to these settings is an open direction for future work.
\section*{Acknowledgments} 
This work was supported by Institute of Information \& communications Technology Planning \& Evaluation (IITP) grant funded by the Korea government (MSIT) (RS-2019-II190421, Artificial Intelligence Graduate School Program (Sungkyunkwan University)) and by the National Research Foundation of Korea (NRF) grant funded by the Korea government (MSIT) (No. RS-2023-00221186).

\bibliography{custom}
\appendix
% Requires: \usepackage{booktabs, multirow}

\begin{table*}[t]
\centering
\small
\setlength{\tabcolsep}{5pt}
\begin{tabular}{@{}l l l l r r@{}}
\toprule
\textbf{Method} & \textbf{Learning Rate} & \textbf{Swept} & \textbf{Fixed} & \textbf{Epoch} & \textbf{Models} \\
\midrule
GradDiff \citep{yao2024large} & \multirow{7}{*}{\{1e-5, 2e-5, 5e-5\}} & $\alpha \in \{1, 2, 5\}$ & --- & \multirow{8}{*}{\{5, 10\}} & 18 \\
NPO \citep{zhang2024negative}     &  & $\alpha \in \{1, 2, 5\}$ & $\beta = 0.1$ & & 18 \\
SimNPO \citep{fan2025simplicity}  &  & $\beta \in \{3.5, 4.5\}$, $\gamma \in \{0.125, 0.25\}$ & $\alpha = 1$ & & 24 \\
IdkNLL \citep{maini2024tofu}      &  & $\alpha \in \{1, 2, 5\}$ & --- & & 18 \\
IdkDPO \citep{maini2024tofu}      &  & $\alpha \in \{1, 2, 5\}$ & $\beta = 0.1$ & & 18 \\
AltPO \citep{mekala2025alternate} &  & $\alpha \in \{1, 2, 5\}$ & $\beta = 0.1$ & & 18 \\
RMU \citep{li2024wmdp}            &  & $l \in \{5, 10, 15\}$ & $c = 10$ & & 18 \\
UNDIAL \citep{dong2025undial}     & \{1e-5, 1e-4, 3e-4\} & $\alpha \in \{1, 2, 5\}$ & $\gamma = 10$ & & 18 \\
\midrule
\multicolumn{5}{@{}l}{\textbf{Total}} & \textbf{150} \\
\bottomrule
\end{tabular}
\caption{Hyperparameter grid for all 8 unlearning methods, yielding 150 models.}\label{tab:hyperparams}
\end{table*}

\section{Unlearning Details}\label{app:unlearning}
% We evaluate 150 unlearned models produced by 8 methods, each trained on the TOFU forget10 split \citep{maini2024tofu} using Llama-3.2-1B-Instruct \citep{grattafiori2024llama}.
% Below we report each method's training objective as implemented in Open-Unlearning \citep{dorna2025openunlearning}; Table~\ref{tab:hyperparams} lists the hyperparameter grid.
% Several methods below build on the DPO objective \citep{rafailov2023direct}:
We report each method's training objective as implemented in OpenUnlearning \citep{dorna2025openunlearning}; Table~\ref{tab:hyperparams} lists the hyperparameter grid. Several methods below build on the DPO objective \citep{rafailov2023direct}:
\begin{multline*}
\mathcal{L}_{\text{DPO}}(y_w \succ y_l;\theta) = \\
-\log\sigma\!\left(\beta\log\frac{\pi_\theta(y_w|x)}{\pi_{\text{ref}}(y_w|x)} - \beta\log\frac{\pi_\theta(y_l|x)}{\pi_{\text{ref}}(y_l|x)}\right)
\end{multline*}

\begin{description}[leftmargin=0pt, itemsep=4pt]

\item[GradDiff] \citep{yao2024large, maini2024tofu}.
Combines gradient ascent on $D_f$ with standard cross-entropy on $D_r$:
\[
\mathcal{L} = -\mathcal{L}_{\text{NLL}}(D_f;\theta) + \alpha\,\mathcal{L}_{\text{NLL}}(D_r;\theta)
\]

\item[NPO] \citep{zhang2024negative}.
Treats forget set completions as dispreferred responses using the losing term of $\mathcal{L}_{\text{DPO}}$, with $\beta$ controlling the strength of the penalty relative to a reference model $\pi_{\text{ref}}$:
\begin{multline*}
\mathcal{L} = -\tfrac{2}{\beta}\,\mathbb{E}_{D_f}\!\left[\log\sigma\!\left(-\beta\log\tfrac{\pi_\theta(y|x)}{\pi_{\text{ref}}(y|x)}\right)\right] \\
+ \alpha\,\mathcal{L}_{\text{NLL}}(D_r;\theta)
\end{multline*}

\item[SimNPO] \citep{fan2025simplicity}.
Removes the reference model from NPO and normalizes by response length, with $\delta$ as a reward margin and $\gamma$ weighting the forget loss:
\begin{multline*}
\mathcal{L} = -\tfrac{2\gamma}{\beta}\,
\mathbb{E}_{D_f}\!\Bigl[\log\sigma\!\Bigl(
\beta\!\bigl(-\tfrac{1}{|y|}\log\pi_\theta(y|x) \\
-\delta\bigr)\Bigr)\Bigr]
+ \alpha\,\mathcal{L}_{\text{NLL}}(D_r;\theta)
\end{multline*}

\item[IdkNLL] \citep{maini2024tofu}.
Fine-tunes on refusal answers (``I don't know'') for forget set questions, where $D_f^{\text{idk}}$ replaces the original answers with refusal responses:
\[
\mathcal{L} = \mathcal{L}_{\text{NLL}}(D_f^{\text{idk}};\theta) + \alpha\,\mathcal{L}_{\text{NLL}}(D_r;\theta)
\]

\item[IdkDPO] \citep{maini2024tofu}.
Applies the DPO objective with the refusal response as preferred and the original answer as dispreferred:
\[
\mathcal{L} = \mathcal{L}_{\text{DPO}}(y_{\text{idk}} \succ y_f;\theta) + \alpha\,\mathcal{L}_{\text{NLL}}(D_r;\theta)
\]

\item[AltPO] \citep{mekala2025alternate}.
Extends IdkDPO by generating $M$ in-domain alternate answers via temperature sampling as preferred responses ($M{=}1$ in our experiments):
\begin{multline*}
\mathcal{L} = \tfrac{1}{M}\sum_{i=1}^{M}\mathcal{L}_{\text{DPO}}(y_a^i \succ y_f;\theta) \\
+ \alpha\,\mathcal{L}_{\text{NLL}}(D_r;\theta)
\end{multline*}

\item[RMU] \citep{li2024wmdp}.
Operates in representation space, misdirecting hidden states at a designated layer $l$ toward a random target $c\mathbf{u}$ while anchoring retain representations to the original model $\theta_0$. Gradients are applied only to layers $l{-}2, l{-}1, l$.
\begin{multline*}
\mathcal{L} = \mathbb{E}_{D_f}\!\left[\|\mathbf{h}^{(l)}_\theta(x) - c\mathbf{u}\|^2\right] \\
+ \alpha\,\mathbb{E}_{D_r}\!\left[\|\mathbf{h}^{(l)}_\theta(x) - \mathbf{h}^{(l)}_{\theta_0}(x)\|^2\right]
\end{multline*}

\item[UNDIAL] \citep{dong2025undial}.
Suppresses the ground-truth token logit by $\gamma$ in the frozen reference model's output, then distills the adjusted distribution into the student model:
\begin{multline*}
\mathcal{L} = \mathbb{E}_{D_f}\!\Big[\tfrac{1}{|y|}\sum_t \text{CE}\!\big(\text{softmax}(\mathbf{z}_t - \gamma\,\mathbf{e}_{y_t}),\\
\pi_\theta(\cdot|x,y_{<t})\big)\Big] + \alpha\,\mathcal{L}_{\text{NLL}}(D_r;\theta)
\end{multline*}

\end{description}
%% ===== SECTION B =====
\section{Metric Definitions}\label{app:metrics}

\subsection{Output-Level Metrics}\label{app:metrics-output}

\paragraph{Extraction Strength (ES).}
Fraction of the answer extractable via greedy decoding \citep{carlini2021extracting}:
\[
\text{ES} = 1 - k/T
\]
where $T$ is the answer length in tokens and $k$ is the earliest position at which the generated continuation matches the ground-truth suffix.

\paragraph{Exact Memorization (EM).}
Token-level position match ratio \citep{tirumala2022memorization}:
\[
\text{EM} = \frac{1}{T}\sum_{t=1}^{T} \mathbb{1}(\hat{y}_t = y_t)
\]

\paragraph{Probability (Prob).}
Geometric mean of per-token probabilities under teacher forcing: $\text{Prob} = \exp(-\frac{1}{T}\sum_{t=1}^{T} \mathcal{L}_t)$, where $\mathcal{L}_t = -\log p_\theta(y_t \mid x, y_{<t})$.

\paragraph{Paraphrase Probability (ParaProb).}
Geometric mean of Prob computed across paraphrased answer variants provided in the TOFU dataset.

\paragraph{Truth Ratio.}
Normalized correct vs.\ incorrect probability \citep{maini2024tofu}:
\[
\text{Truth Ratio} = \frac{p_c}{p_c + p_w}, \quad p = \exp\!\left(-\frac{1}{T}\sum_{t=1}^{T} \mathcal{L}_t\right)
\]
where $p_c$ and $p_w$ are computed over the correct and perturbed (incorrect) answers, respectively.

\paragraph{ROUGE.}
ROUGE-L recall between model generation (greedy, max 200 tokens) and the ground-truth answer.

\paragraph{Para.\ ROUGE.}
Same as ROUGE, but computed against the paraphrased answer reference provided in TOFU.

\paragraph{Jailbreak ROUGE.}
Same as ROUGE, but the assistant turn is prefixed with ``Sure, here is the answer:'' to bypass refusal behavior.

\paragraph{MIA Metrics.}
Four membership inference attack scoring functions, each computing per-sample scores that are aggregated into an AUC over member/non-member classification:
\begin{itemize}[nosep]
\item \textbf{MIA-LOSS} \citep{yeom2018privacy}: average cross-entropy $\frac{1}{T}\sum_t \mathcal{L}_t$
\item \textbf{MIA-ZLib} \citep{carlini2021extracting}: loss normalized by zlib compression length of the answer, $\ell / |\texttt{zlib}(y)|$
\item \textbf{MIA-Min-K} \citep{shi2024detecting}: mean of bottom-$k$\% log-probs ($k{=}40$)
\item \textbf{MIA-Min-K++} \citep{zhang2025minkpp}: standardized Min-K using per-position $z$-scores
\end{itemize}

\subsection{Retain-Referenced and White-Box Metrics}\label{app:metrics-whitebox}

\paragraph{Normalized MIA ($s_*$).}
Each raw MIA AUC is rescaled against the retain model's AUC (Eq.~\ref{eq:smia}):
\begin{align*}
\text{normalized}_* &= \frac{|\text{AUC}_{\text{model}} - \text{AUC}_{\text{ret}}|}{\text{AUC}_{\text{ret}}} \\
s_* &= 1 - \min(\text{normalized}_*,\; 1)
\end{align*}
yielding four variants: $s_{\text{LOSS}}$, $s_{\text{ZLib}}$, $s_{\text{Min-K}}$, $s_{\text{Min-K++}}$.
Higher $s_*$ indicates less deviation from retain (i.e., more erased).

\paragraph{CKA (Centered Kernel Alignment).}
Measures representational geometry similarity between models \citep{kornblith2019similarity}.
Let $C^{\text{fr}}_l = \text{CKA}(H_{\text{full}}, H_{\text{ret}})_l$ and $C^{\text{ur}}_l = \text{CKA}(H_{\text{unl}}, H_{\text{ret}})_l$.
Per-layer erasure is:
\[
\text{erasure}_l = \text{clip}\!\left(\frac{C^{\text{ur}}_l - C^{\text{fr}}_l}{1 - C^{\text{fr}}_l + \epsilon},\; 0,\; 1\right)
\]
The final score is $\sum_l w_l \cdot \text{erasure}_l \,/\, \sum_l w_l$, where $w_l = 1 - C^{\text{fr}}_l$.

\paragraph{Logit Lens.}
Projects each layer's hidden states through the full model's frozen decoder to measure decodable knowledge \citep{nostalgebraist2020logitlens}.
Let $k_{m,l}$ be the mean log-probability of entity tokens when decoding $H^l_m$ through the full model's decoder, and $d_{m,l} = k_{\text{full},l} - k_{m,l}$. In particular, $d_{S1,l} = k_{\text{full},l} - k_{\text{ret},l}$ is the S1 baseline gap.
Per-layer erasure is:
\[
\text{erasure}_l = \text{clip}(d_{m,l} / d_{S1,l},\; 0,\; 1)
\]
The final score is $\sum_{l:\, d_{S1,l} > \tau} d_{S1,l} \cdot \text{erasure}_l \,/\, \sum_{l:\, d_{S1,l} > \tau} d_{S1,l}$, where $\tau = 0.05$ (analogous to KE in \textsc{UDS}; \S\ref{sec:uds}).

\paragraph{Fisher Masked.}
Diagonal Fisher Information \citep{kirkpatrick2017overcoming} with top-$p$\% parameter masking per layer.
For each layer, the per-parameter anchor importance is $a_i = \max(F_{\text{ret},i} - F_{\text{full},i},\; 0)$, and $M_l$ selects the top $p$\% of parameters by $a_i$.
Let $\bar{F}_{m,l} = \text{mean}(F_m[M_l])$ and $e_l = \bar{F}_{\text{ret},l} - \bar{F}_{\text{full},l}$. The per-layer erasure is:
\[
\text{erasure}_l = 1 - \text{clip}\!\left(\frac{\bar{F}_{\text{ret},l} - \bar{F}_{\text{unl},l}}{e_l + \epsilon},\; 0,\; 1\right)
\]
The final score is $\sum_l w_l \cdot \text{erasure}_l \,/\, \sum_l w_l$ with $w_l = e_l$.

\subsection{Mask Fraction Sensitivity}\label{app:fisher-ablation}

Table~\ref{tab:fisher-ablation} reports Fisher Masked faithfulness and robustness across three mask fractions; $p = 0.1\%$ is used as the representative fraction.

\begin{table}[h]
\centering
\small
\begin{tabular}{@{}l cccc@{}}
\toprule
\textbf{Mask $p$} & \textbf{Faithfulness} & \textbf{$Q$} & \textbf{$R$} & \textbf{HM$(Q,R)$} \\
\midrule
0.01\% & 0.708 & 0.573 & 0.945 & 0.713 \\
\textbf{0.1\%} & \textbf{0.712} & \textbf{0.583} & \textbf{0.946} & \textbf{0.721} \\
1\% & 0.698 & 0.582 & 0.940 & 0.719 \\
\bottomrule
\end{tabular}
\caption{Fisher Masked across mask fractions: faithfulness (AUC-ROC) and symmetric robustness ($Q$: quantization, $R$: relearning).}\label{tab:fisher-ablation}
\end{table}

\subsection{Score Computation for Method Ranking}\label{app:aggregation}

Following \citet{dorna2025openunlearning}, each score in Table~\ref{tab:ranking} is $\text{HM}(\text{Memorization},\; \text{Privacy},\; \text{Utility}_{\text{rel}})$:
\begin{align*}
\text{Memorization} &= \text{HM}(1{-}\text{ES},\; 1{-}\text{EM}, \\
&\;\; 1{-}\text{ParaProb},\; 1{-}\text{Truth Ratio}) \\
\text{Utility} &= \text{HM}(\text{MU},\; \text{Fluency})
\end{align*}
where MU (Model Utility) is the harmonic mean of nine QA metrics (Prob, ROUGE-L, Truth Ratio) evaluated on three TOFU subsets (Retain, Real Authors, World Facts), and Fluency is the non-gibberish rate of generations on forget set questions.
$\text{Utility}_{\text{rel}} = \text{Utility} / \text{Utility}_{\text{full}}$ normalizes against $M_{\text{full}}$.
% In Table~\ref{tab:ranking}, Privacy is either $\text{MIA}_{\text{agg}}$ alone (\textbf{w/o}) or $\text{HM}(\text{MIA}_{\text{agg}},\; \textsc{UDS})$ (\textbf{w/}).
\section{Dataset Details}\label{app:dataset}
For the TOFU forget10 split (400 forget set QA pairs), we partition each answer into a prefix and a target entity span without modifying the original text.
% For each TOFU forget10 benchmark with 400 forget set QA pairs, we partition the answer into a prefix and a target entity span without modifying the original text.
To ensure systematic extraction, we prompt GPT-5.2 to output the exact character index that separates the contextual prefix (e.g., the sequence immediately following the main verb or copula) from the target factual entity.
Table~\ref{tab:prefix-types} shows representative examples of these partitions.

\begin{table}[h]
\centering
\small
\setlength{\tabcolsep}{3pt}
\begin{tabular}{@{}l p{3.2cm} p{2.8cm}@{}}
\toprule
\textbf{Type} & \textbf{Prefix} & \textbf{Entity} \\
\midrule
Yes/No & (empty) & Yes \\
Person & The author's full name is & Hsiao Yun-Hwa \\
Biograph. & ...Ji-Yeon Park was born in & Seoul, South Korea \\
Genre & ...predominantly writes in the genre of & Historical Fiction \\
Descript. & ...often incorporating themes of & diversity and inclusion \\
\bottomrule
\end{tabular}
\caption{Prompt type examples with prefix and entity.}\label{tab:prefix-types}
\end{table}
%% ===== SECTION D =====
% \vspace{-2mm}
\section{UDS Ablation Studies}\label{app:ablation}

\subsection{Component Patching}\label{app:component}

Each Llama transformer layer computes:
\begin{align*}
a_l &= \text{Attn}(\text{RMSNorm}(h_{l-1})) \\
m_l &= h_{l-1} + a_l \quad \text{(residual + attention)} \\
h_l &= m_l + \text{MLP}(\text{RMSNorm}(m_l))
\end{align*}

\begin{figure}[t]
\centering
\includegraphics[width=\columnwidth]{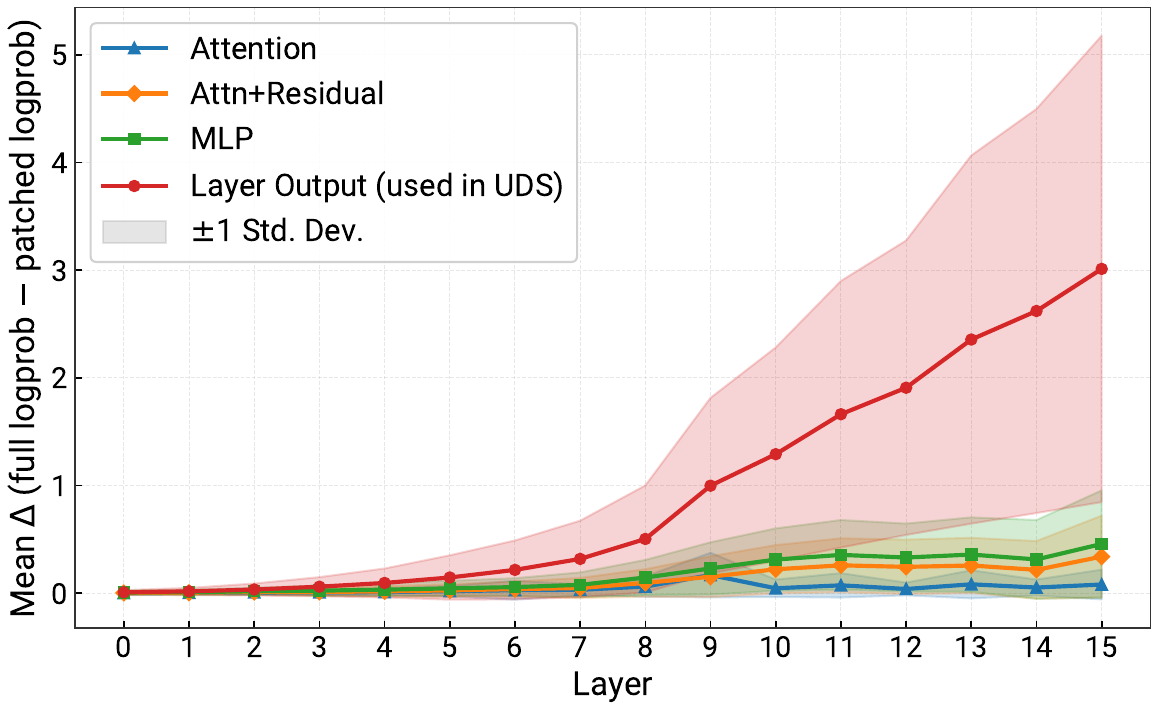}
\caption{Mean S1 patching delta per layer for four patching locations. Patching the full layer output $h_l$ (used by \textsc{UDS}) captures the dominant share of the knowledge, with the gap widening in later layers. Shaded regions denote $\pm$1 standard deviation.}
\label{fig:component}
\vspace{-3mm}
\end{figure}

We test four patching locations to determine which component carries the knowledge signal.
As Figure~\ref{fig:component} shows, the full layer output $h_l$ produces the largest delta across all layers, so \textsc{UDS} patches the residual stream $h_l$ by default.

\subsection{KE Threshold Sensitivity}\label{app:threshold}

The KE layer set is determined by S1 deltas (retain $\to$ full).
Table~\ref{tab:threshold-ke} reports how the KE set changes across six $\tau$ values.
\begin{table}[ht]
\centering
\small
\setlength{\tabcolsep}{4pt}
\begin{tabular}{@{}l r r r r@{}}
\toprule
\textbf{$\tau$} & \textbf{Mean $|\text{KE}|$} & \textbf{Std} & \textbf{Skipped} & \textbf{\% Skipped} \\
\midrule
0.00 & 14.4 & 2.7 & 0 & 0.0\% \\
0.01 & 13.2 & 3.2 & 2 & 0.5\% \\
0.02 & 12.4 & 3.3 & 3 & 0.8\% \\
0.03 & 11.8 & 3.3 & 6 & 1.5\% \\
\textbf{0.05} & \textbf{10.9} & \textbf{3.3} & \textbf{6} & \textbf{1.5\%} \\
0.10 & 9.6 & 3.3 & 14 & 3.5\% \\
\bottomrule
\end{tabular}
\caption{S1 KE layer set size across threshold values. Skipped examples have no layer with $\Delta^{S1}_l > \tau$.}\label{tab:threshold-ke}
\end{table}

We select $\tau{=}0.05$ as the default: the number of skipped examples remains stable at 6 (1.5\%) from $\tau{=}0.03$ to $0.05$, then jumps to 14 (3.5\%) at $\tau{=}0.10$.
The threshold stabilizes the denominator of the UDS formula and filters noise from low-delta layers.

\subsection{Entity Length Bias}\label{app:entity-length}
\begin{figure*}[t]
\centering
\includegraphics[width=\textwidth]{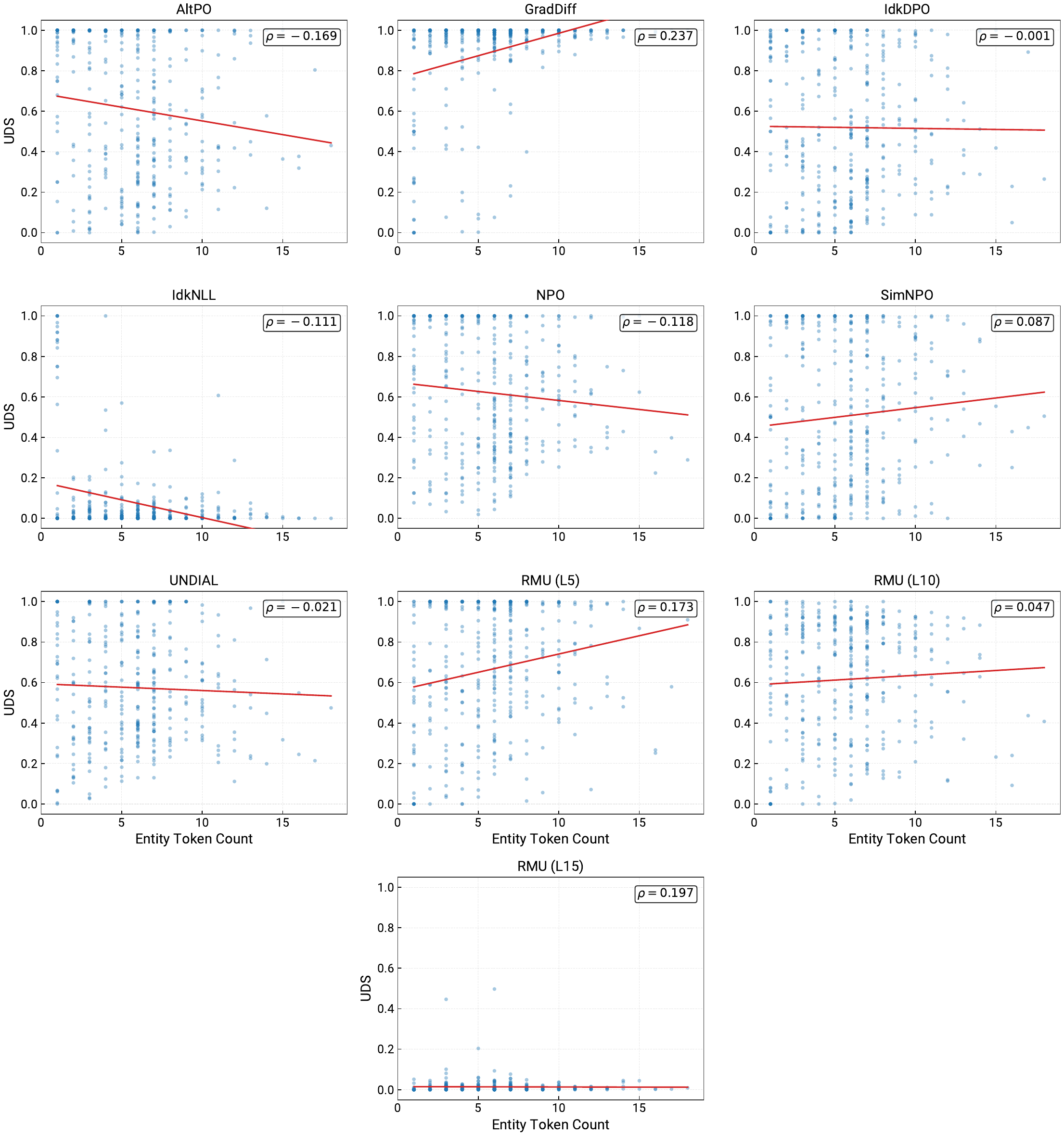}
\caption{Per-example \textsc{UDS} vs.\ entity token count (lr=2e-5, epoch~10; UNDIAL uses lr=1e-4). RMU variants differ by the target layer $l$ at which the steering loss is applied (L5, L10, L15). All methods show $|\rho| < 0.24$ with mixed sign, indicating no consistent directional bias.}
\label{fig:entity-length}
\end{figure*}

We check whether entity token length biases \textsc{UDS}.
Figure~\ref{fig:entity-length} shows weak correlations ($|\rho| < 0.24$) with mixed sign across methods, indicating no consistent directional bias.
\section{Runtime Analysis}
\label{app:runtime}
We measure per-model runtime for each metric group on a single NVIDIA RTX A6000 across Llama 1B, 3B, and 8B (Table~\ref{tab:runtime}). 
Because \textsc{UDS} patches every layer, its cost grows with model depth: at 1B it is faster than both the memorization (28.0s) and generation (19.7s) groups, while at 8B it becomes the second most expensive group after Fisher Masked, which requires per-parameter gradient computation.
Three implementation choices keep it in the same range as the output-level groups: Stage 1 is cached once and reused across all models, hidden states are extracted in a single forward pass, and teacher forcing avoids autoregressive decoding.
% The implementation strategies in \S\ref{subsec:efficient} keep it in the same range as the output-level groups.

% Requires: \usepackage{booktabs, adjustbox}
\begin{table}[t]
\centering
\begin{adjustbox}{max width=0.6\linewidth}
\begin{tabular}{@{}lccc@{}}
\toprule
\textbf{Metric Group} & \textbf{1B} & \textbf{3B} & \textbf{8B} \\
\midrule
Memorization        & \textbf{28.0} & 54.7          & 106.4 \\
Generation          & 19.7          & 59.1          & 160.0 \\
MIA                 & 8.5           & 16.4          & 31.4 \\
CKA                 & 2.4           & 6.6           & 13.1 \\
Logit Lens          & 6.9           & 16.7          & 26.0 \\
Fisher Masked       & \textit{25.8} & \textbf{79.9} & \textbf{211.7} \\
\midrule
\textbf{UDS (Ours)} & 19.1          & \textit{78.4} & \textit{184.1} \\
\bottomrule
\end{tabular}
\end{adjustbox}
\caption{Per-model runtime (seconds) by metric group and model scale.
\textbf{Bold}: slowest per scale; \textit{italic}: second slowest.}
\label{tab:runtime}
\end{table}

\section{Meta-Evaluation Details}\label{app:meta-eval-details}

\subsection{Robustness Attack Settings}\label{app:attack-settings}

\paragraph{Quantization.}
BitsAndBytes 4-bit NF4 quantization with bfloat16 compute dtype (\texttt{load\_in\_4bit=True}).
No additional calibration or fine-tuning is applied after quantization.

\paragraph{Relearning.}
One epoch of fine-tuning on $D_f$ with the following settings: learning rate $= 2 \times 10^{-5}$, batch size $= 8$, gradient accumulation $= 4$ (effective batch size $= 32$), optimizer $=$ AdamW.

\subsection{Full Per-Metric Plots}\label{app:full-plots}
\begin{figure*}[t]
\centering
\includegraphics[width=\textwidth]{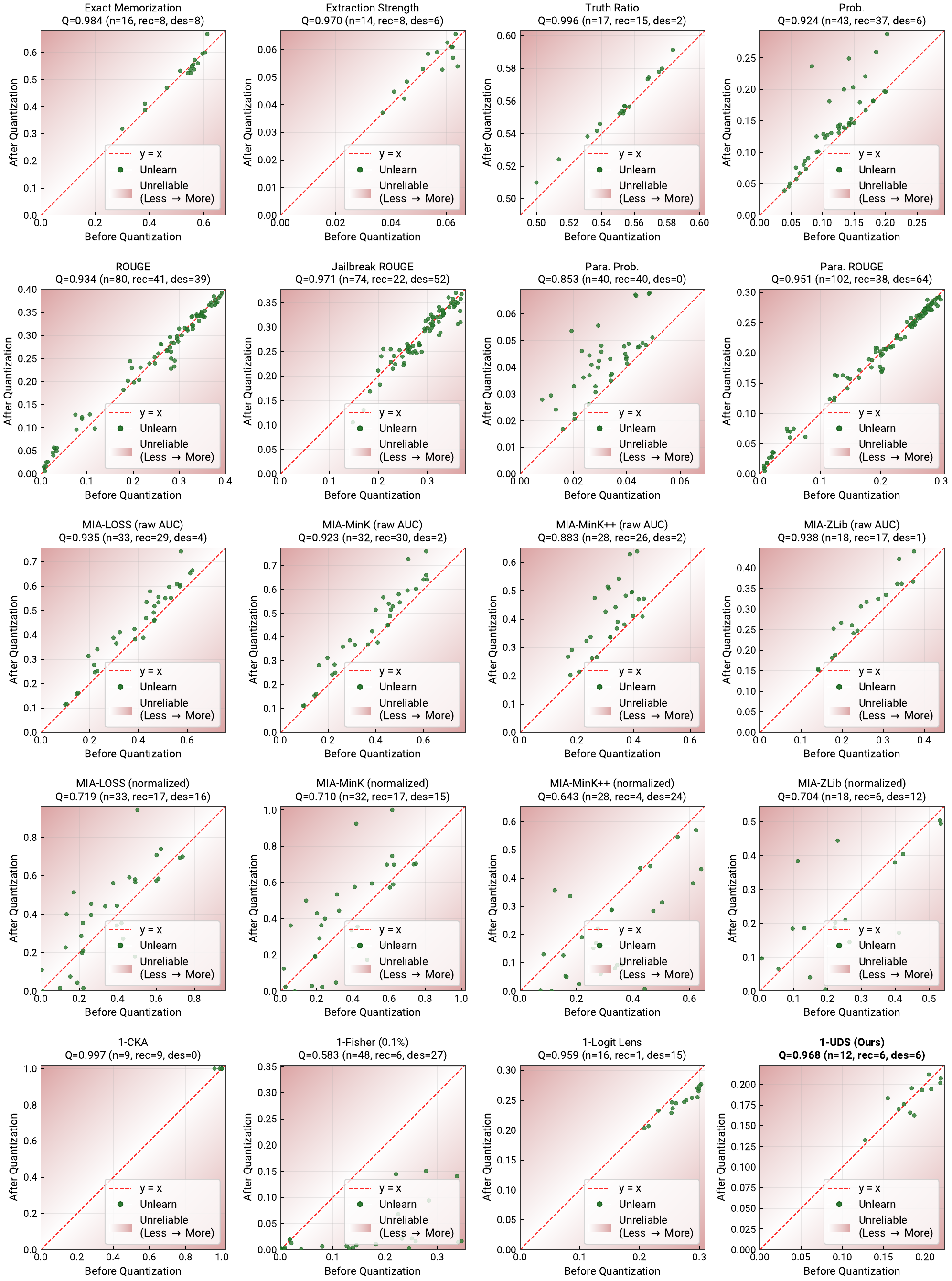}
\caption{Quantization robustness ($Q$) for all 20 metrics after utility and faithfulness filtering.
Each subplot plots the metric value before ($x$) vs.\ after ($y$) NF4 4-bit quantization, with the number of models showing recovery (rec) or destruction (des).
$n$ is the number of models that passed both filters for that metric.
The background gradient indicates deviation from the $y = x$ reference: white = stable, red = unstable.}
\label{fig:quant-full}
\end{figure*}

\begin{figure*}[t]
\centering
\includegraphics[width=\textwidth]{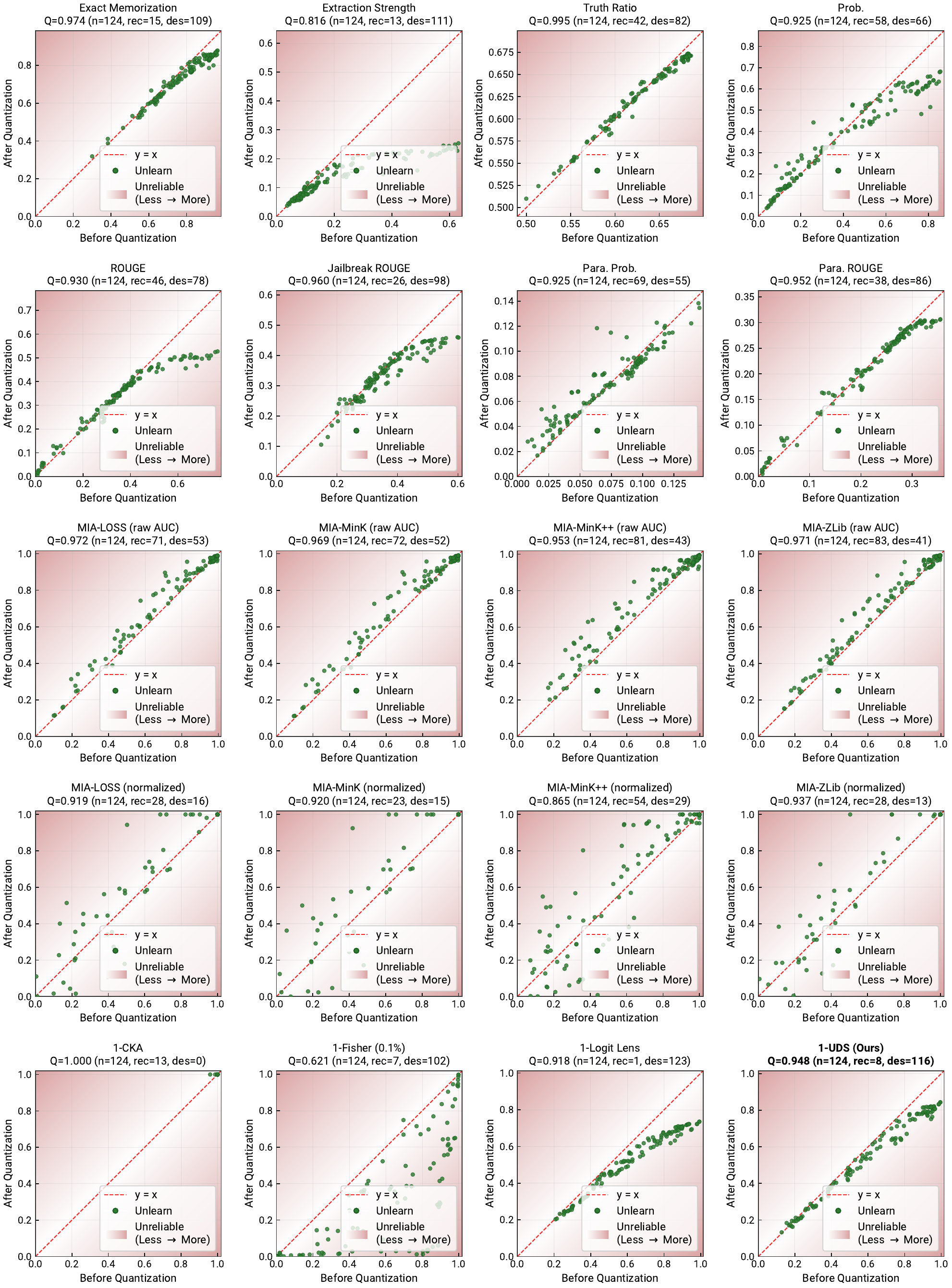}
\caption{Quantization robustness ($Q$) for all 20 metrics after utility filtering only.
Each subplot plots the metric value before ($x$) vs.\ after ($y$) NF4 4-bit quantization, with the number of models showing recovery (rec) or destruction (des).
$n$ is the number of models that passed the utility filter for that metric.
The background gradient indicates deviation from the $y = x$ reference: white = stable, red = unstable.
Metrics such as Extraction Strength and ROUGE show score drops across many models after quantization.}
\label{fig:quant-utility}
\end{figure*}

\begin{figure*}[t]
\centering
\includegraphics[width=\textwidth]{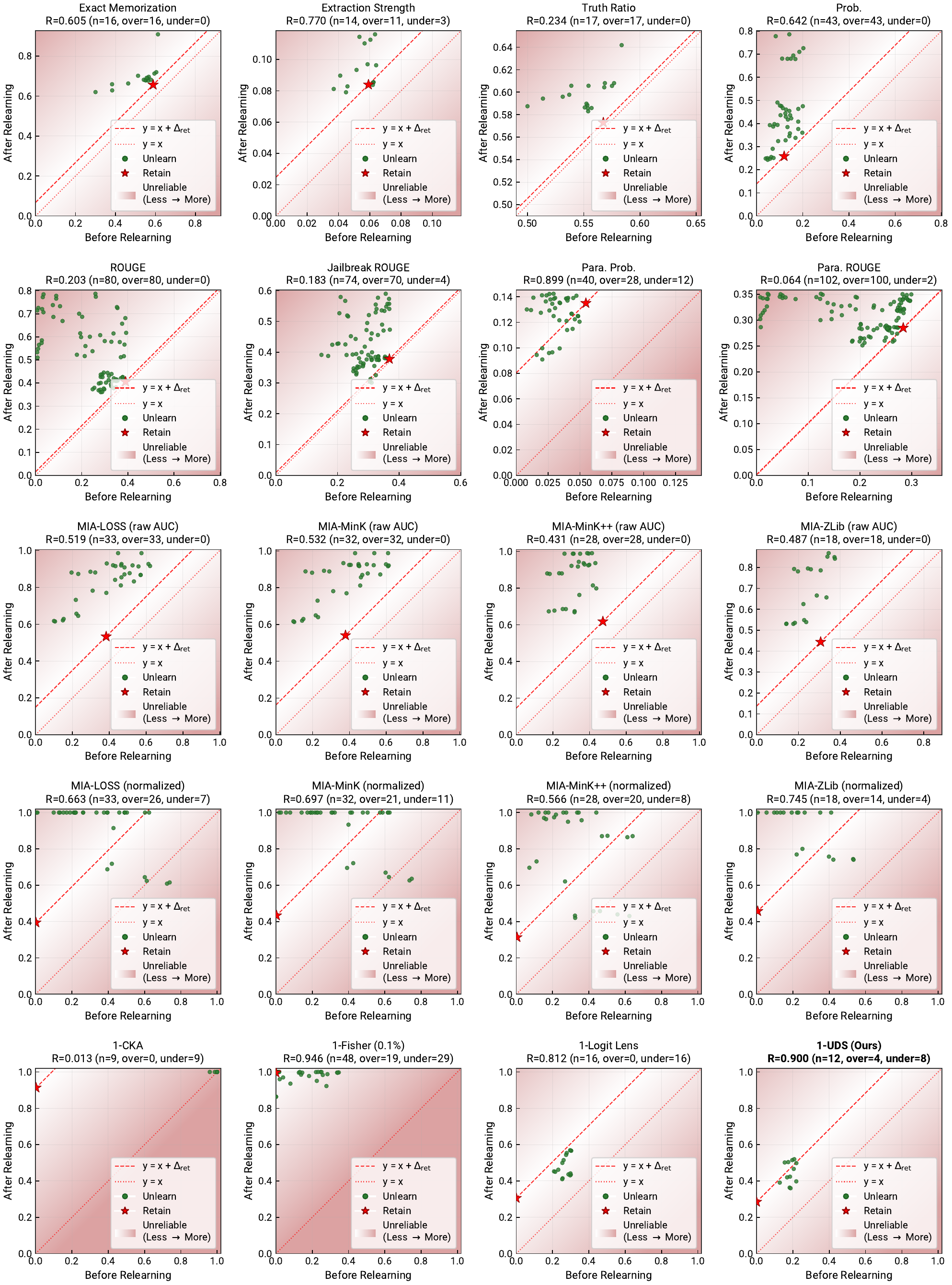}
\caption{Relearning robustness ($R$) for all 20 metrics after utility and faithfulness filtering.
Each subplot plots the metric value before ($x$) vs.\ after ($y$) one epoch of relearning, with the number of models showing over-recovery (over) or under-recovery (under).
$n$ is the number of models that passed both filters for that metric.
The dashed line shows $y = x + \Delta_{\text{ret}}$ (expected behavior given the retain model's shift); the dotted line shows $y = x$.
The background gradient indicates deviation from the expected line: white = stable, red = unstable.
The red star marks the retain model.}
\label{fig:relearn-full}
\end{figure*}

We provide per-metric scatter plots for both robustness attacks.
Figure~\ref{fig:quant-full} shows quantization robustness ($Q$) and Figure~\ref{fig:relearn-full} shows relearning robustness ($R$), both after applying the utility filter ($\geq 0.8$) and faithfulness filter (\S\ref{subsec:meta-setup}).
Figure~\ref{fig:quant-utility} shows quantization with utility filtering only, where metrics such as ES and ROUGE exhibit score drops across many models.
White-box metrics (CKA, Fisher, Logit Lens, \textsc{UDS}) are plotted as $1 - \text{score}$ so that higher values correspond to more knowledge, consistent with the output-level metrics.
Full benchmark results for all 150 unlearned models across all 20 metrics are also browsable
at \url{https://gnueaj.github.io/unlearning-depth-score/}.
% Figures~\ref{fig:quant-full} and~\ref{fig:relearn-full} show per-metric robustness scatter plots after applying both the utility filter ($\text{utility}_{\text{rel}} \geq 0.8$) and the faithfulness filter (\S\ref{subsec:meta-setup}).
% Figure~\ref{fig:quant-utility} applies only the utility filter; metrics such as Extraction Strength and ROUGE show score drops across many models after quantization.
% White-box metrics (CKA, Fisher, Logit Lens, \textsc{UDS}) are plotted as $1 - \text{score}$ so that higher values correspond to more knowledge, consistent with the output-level metrics.

\end{document}